\documentclass[10pt,journal,compsoc]{IEEEtran}

\usepackage{amsmath,amsfonts}
\usepackage{algorithmic}
\usepackage{algorithm}
\usepackage{array}
\usepackage[caption=false,font=normalsize,labelfont=sf,textfont=sf]{subfig}
\usepackage{textcomp}
\usepackage{stfloats}
\usepackage{url}
\usepackage{verbatim}
\usepackage{graphicx}
\usepackage{cite}
\usepackage{svg}
\usepackage{booktabs}
\usepackage{threeparttable} % optional, for notes
\usepackage{siunitx}        % for consistent number formatting (optional)
\usepackage{makecell}       % for multi-line headers
\usepackage{gensymb}

\definecolor{ll_color}{HTML}{7F00FF}
\definecolor{hl_color}{HTML}{E4E45B}
\definecolor{lh_color}{HTML}{FF0080}
\definecolor{hh_color}{HTML}{FF8000}

\begin{document}

\title{WaveMAE: Wavelet decomposition Masked Auto-Encoder for Remote Sensing}

\author{Vittorio Bernuzzi, Leonardo Rossi, Tomaso Fontanini, Massimo Bertozzi, and Andrea Prati \\
%Università degli Studi di Parma\\
%Parma, 43124, IT\\
%{\tt\small \{vittorio.bernuzzi,leonardo.rossi,tomaso.fontanini,massimo.bertozzi,andrea.prati\}@unipr.it}
        % <-this % stops a space
\thanks{Vittorio Bernuzzi, Leonardo Rossi, Tomaso Fontanini, Massimo Bertozzi, and Andrea Prati are with the Dipartimento di Ingegneria e Architettura of the Università di Parma (Parma, I-43124, Italy) (e-mail: \{vittorio.bernuzzi, leonardo.rossi, tomaso.fontanini, massimo.bertozzi, andrea.prati\}@unipr.it)}% <-this % stops a space
\thanks{The code will be released at the Github repository IMPLabUniPr.}}

\markboth{}%
{V. Bernuzzi \MakeLowercase{\textit{et al.}}: WaveMAE: Wavelet decomposition Masked Auto-Encoder for Remote Sensing}

\maketitle

\begin{abstract}
Self-supervised learning (SSL) has recently emerged as a key strategy for building foundation models in remote sensing, where the scarcity of annotated data limits the applicability of fully supervised approaches.
In this work, we introduce \textbf{WaveMAE}, a masked autoencoding framework tailored for multispectral satellite imagery.
Unlike conventional pixel-based reconstruction, WaveMAE leverages a multi-level \textit{Discrete Wavelet Transform} (DWT) to disentangle frequency components and guide the encoder toward learning scale-aware high-frequency representations.
We further propose a \textit{Geo-conditioned Positional Encoding} (GPE), which incorporates geographical priors via Spherical Harmonics, encouraging embeddings that respect both semantic and geospatial structure.
To ensure fairness in evaluation, all methods are pretrained on the same dataset (fMoW-S2) and systematically evaluated on the diverse downstream tasks of the \textsc{PANGAEA} benchmark, spanning semantic segmentation, regression, change detection, and multilabel classification.
Extensive experiments demonstrate that WaveMAE achieves consistent improvements over prior state-of-the-art approaches, with substantial gains on segmentation and regression benchmarks.
The effectiveness of WaveMAE pretraining is further demonstrated by showing that even a lightweight variant, containing only $26.4$\% of the parameters, achieves state-of-the-art performance.
Our results establish WaveMAE as a strong and geographically informed foundation model for multispectral remote sensing imagery.
\end{abstract}


\section{Introduction}
\label{sec:intro}

\IEEEPARstart{I}{n} recent years, self-supervised learning (SSL) has emerged as a powerful paradigm for pretraining deep neural networks on Remote Sensing (RS) data, enabling the extraction of meaningful representations without relying on costly human annotations \cite{survey_ssl}. 
This is particularly valuable for satellite imagery, where large-scale unlabeled archives are abundant but high-quality labeled datasets remain scarce and task-specific annotations are expensive to obtain. 
By leveraging intrinsic structures in the data, SSL alleviates the need for manual labels and provides transferable representations across a wide range of downstream applications.

Multispectral satellite imagery, exemplified by the Sentinel-2 mission, introduces unique challenges and opportunities for representation learning. 
Sentinel-2 offers 13 spectral bands ranging from visible and near-infrared (VNIR) to short-wave infrared (SWIR) at spatial resolutions of 10, 20, and 60 meters \cite{sentinel2_paper}. 
This rich spectral diversity supports fine-grained analysis of land cover, vegetation health, and environmental monitoring, but also demands feature encoders that can preserve spectral coherence while effectively capturing semantic structure across scales. 
Unlike natural images, RS imagery is characterized by high dimensionality, spectral redundancy, spatial autocorrelation, and temporal dynamics, which complicate direct adoption of standard SSL methods. 
Among the diverse SSL paradigms, masked image modeling (MIM) has recently proven particularly effective for RS data. 
Inspired by MIM, Masked Autoencoders (MAE) \cite{mae_paper} leverage partial reconstruction of the input as a pretext task during pretraining, encouraging models to capture both spatial and spectral dependencies. 
This approach has been extended to remote sensing with domain-specific adaptations, such as SatMAE \cite{satmae_paper}, SatMAE++ \cite{satmae++_paper}, and SpectralGPT \cite{spectralgpt_paper}, all of which emphasize the need for spectral-aware pretraining strategies. 
Nevertheless, existing methods predominantly reconstruct images at the pixel level, potentially limiting their ability to disentangle spectral-frequency structures that are critical for multispectral imagery.

To address this limitation, we introduce the use of the Discrete Wavelet Transform (DWT) as a decomposition strategy within masked autoencoding. 
By explicitly separating frequency components across multiple resolutions, DWT provides a natural inductive bias for capturing both low-frequency spatial context and high-frequency spectral details, thereby improving representation quality and reconstruction fidelity. 
In addition, we propose a novel Geo-conditioned Positional Encoding (GPE), which incorporates geolocation priors via Spherical Harmonics (SH) into the Transformer encoder backbone. 
This injects geographical structure into the learned embeddings, aligning representations according to spatial proximity while preserving semantic information. 
We validate our approach through extensive experiments on the recently proposed \textsc{PANGAEA}-bench \cite{pangaea_paper}, a unified benchmark for SSL in RS that enables fair comparison across methods and downstream tasks. 
Our results demonstrate that the proposed WaveMAE consistently outperforms prior foundation models on a diverse suite of datasets, spanning semantic segmentation, regression, and multilabel classification.

Our main contributions are summarized as follows:
\begin{itemize}
    \item We propose WaveMAE, a novel self-supervised masked autoencoding framework for multispectral remote sensing data that leverages DWT decomposition to disentangle spatial and spectral-frequency components, producing richer representations. 
    \item We introduce a new Geo-conditioned Positional Encoding (GPE) that embeds geolocation information via spherical harmonics into the transformer encoder, enriching the output features with geographical location priors.
\end{itemize}
\section{Related Work}
\label{sec:related}

\subsection{Masked Image Modeling}
Masked Autoencoder (MAE) architectures have demonstrated remarkable success in computer vision by learning robust representations through the reconstruction of masked image patches \cite{mae_paper}. The core principle involves masking random portions of input images and training the model to predict the missing content, therefore learning a rich representation of the data without any label.
Based on this foundation, several specialized approaches have been developed for remote sensing applications. 
SatMAE~\cite{satmae_paper} is one of the first adaptations of the MAE framework to satellite imagery. The novelties introduced to the framework were specifically aimed to the RS domain to comply with the unique features of this type of data. The method addresses the challenge of applying Vision Transformers to multispectral imagery by dividing bands into groups with same spatial resolution and incorporating a spectral-aware attention mechanisms.
ScaleMAE~\cite{scalemae_paper} introduces Ground Sample Distance positional encoding, which augments the learned embeddings with information about the true spatial scale of the image rather than resolution of the pixels. In addition, it employs a Laplacian pyramid-based decoder to reconstruct both low and high-frequency image components on multiple scales.
SatMAE++ \cite{satmae++_paper} builds upon the original SatMAE framework and extends ScaleMAE findings to multispectral imagery by incorporating a multi-scale reconstruction at different image resolutions.
SpectralGPT \cite{spectralgpt_paper} employs a multi-target reconstruction strategy to locally capture spatial-spectral characteristics and spectrally-sequential information, improving the spectral coherence of the reconstructed target.
In contrast to prior approaches, our WaveMAE framework shifts the reconstruction target from raw image pixels to wavelet components. This design enables a fine-grained and hierarchically-structured treatment of high-frequency content, facilitating learning across multiple scales.

\subsection{Discrete Wavelet Transform in computer vision}
The application of DWT to feature extraction has been extensively studied in computer vision. Traditional approaches have demonstrated the effectiveness of wavelet-driven features for texture analysis, edge detection, and image classification \cite{dwt_feature_extraction_paper}. 
Wave-ViT \cite{wavevit_paper} represents a pioneering effort in the direction of unifying wavelet transforms with vision transformers to create a hybrid architecture that uses both the inductive biases of wavelets and the representational power of self-attention mechanisms.
The concept of wavelet-driven masked image modeling has emerged as a promising direction for efficient visual representation learning \cite{wavelet_masked_paper}, incorporating multilevel reconstruction targets, generated by the discrete wavelet transform.
To the best of our knowledge, that is the first work to incorporate a DWT approach to masked image modeling in a self-supervised setup.

Unlike the previous method, which still operate predominantly in the pixel domain, our WaveMAE focuses entirely on wavelet components—both within the network and as the reconstruction target. By using as input the low- and high-frequency components across multiple decomposition scales, the model learns to capture the hierarchical relationships between frequencies at different resolutions.
\section{Theory priors}
\label{sec:theory}

\subsection{Masked Autoencoder}
Masked Autoencoder (MAE) follows the principle that effective visual representations can be learned through large-scale masked reconstruction, leveraging the redundancy in natural images \cite{survey_ssl}. 
The architecture design is built around an asymmetric encoder-decoder network where a high portion of the image patches are randomly corrupted and masked (typically 75\% of the image) before being processed by the encoder \cite{mae_paper}.
After encoding the visible tokens, masked tokens are introduced to complete the sequence and let the decoder reconstruct them. 
This information bottleneck allows the encoder to push high-level semantics into the produced representation  \cite{bottleneck_paper}, allowing the decoder to perform a meaningful reconstruction. 
Moreover, the high masking ratio prevents the model from overfitting low-level pixel statistics, and it encourages learning higher-order visual concepts. Unlike contrastive methods that require carefully designed data augmentations and negative sampling techniques, MAE's reconstruction objective provides a natural learning signal that scales effectively with data size, aligning with the theoretical understanding that generative modeling can yield representations competitive with discriminative approaches.

\subsection{Discrete Wavelet Transform}
\begin{figure}[htbp]
\centerline{\includegraphics[width=\linewidth]{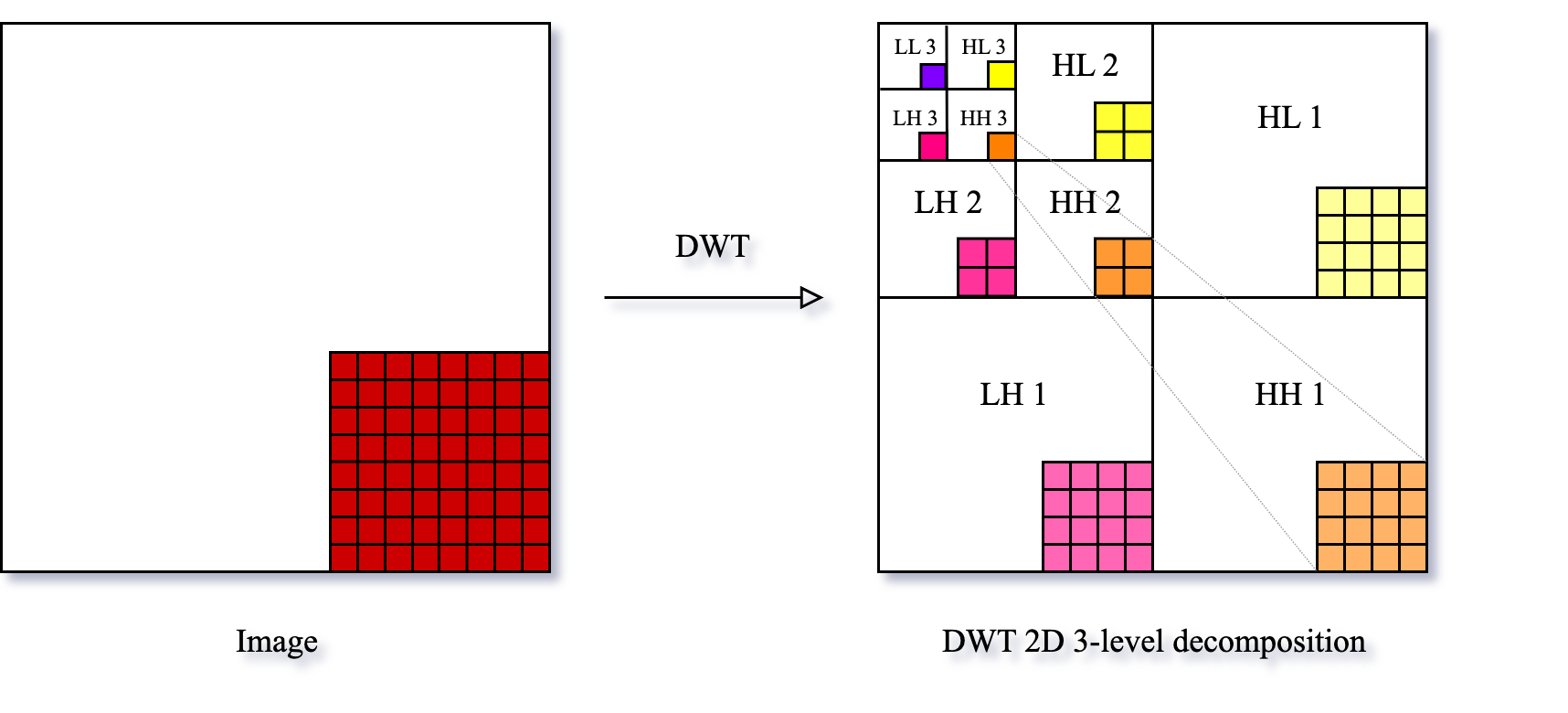}}
\caption{Representation of a 3 level 2D Discrete Wavelet Transform applied on an image. The decomposition from pixel values highlighted in \textcolor{red}{red} generates the wavelet components \textcolor{ll_color}{LL} (low frequencies), \textcolor{lh_color}{LH} (vertical high frequencies), \textcolor{hl_color}{HL} (horizontal high frequencies), and \textcolor{hh_color}{HH} (diagonal high frequencies) on different scales.}
\label{image:2d_dwt}
\end{figure}

The Discrete Wavelet Transform (DWT) is a mathematical framework for analyzing signals in both the time and frequency domains simultaneously. For a one-dimensional discrete signal $x$ of length $N$, the DWT decomposes the signal into approximation components $A_j$ (low-frequency) and detail components $D_j$ (high-frequency) through a series of high-pass and low-pass filtering operations followed by downsampling:
\begin{equation}
    \begin{aligned}
        \text{Approximation: } & \quad A_j[k] = \sum_{n} x[n] \cdot \eta_{j,k}[n] \\
        \text{Detail: } & \quad D_j[k] = \sum_{n} x[n] \cdot \psi_{j,k}[n]
    \end{aligned}
\end{equation}
where $\eta_{j,k}[n]$ and $\psi_{j,k}[n]$ represent the scaling and high-pass filter functions at scale $j$ and translation $k$, respectively.
In the context of image processing, the 2D DWT extends this decomposition to the spatial dimensions. 
The decomposition provides a hierarchical representation of image content on different scales, as illustrated in Figure \ref{image:2d_dwt}.
A single-level 2D Discrete Wavelet Transform decomposition is performed by operating a convolution with a pair of complementary filters $\eta$ (scaling function) and $\psi$ (high-pass filter) with the image $I(x,y)$ and downsamples the latter by a factor of 2 in each dimension:

\begin{equation}
    \begin{aligned}
    \text{LL}_{j+1}(x,y) &= \sum_{m,n} \eta(m)\eta(n) \cdot \text{LL}_j(2x-m, 2y-n) \\
    \text{LH}_{j+1}(x,y) &= \sum_{m,n} \eta(m)\psi(n) \cdot \text{LL}_j(2x-m, 2y-n) \\
    \text{HL}_{j+1}(x,y) &= \sum_{m,n} \psi(m)\eta(n) \cdot \text{LL}_j(2x-m, 2y-n) \\
    \text{HH}_{j+1}(x,y) &= \sum_{m,n} \psi(m)\psi(n) \cdot \text{LL}_j(2x-m, 2y-n)
    \end{aligned}
\end{equation}

\noindent where $\text{LL}_0 = I(x,y)$ is the original image, and $j >= 0$ indicates the decomposition level, $m$ and $n$ are filter coefficient indices. This process yields four components: LL (Low-Low, approximation), LH (Low-High, vertical details), HL (High-Low, horizontal details), and HH (High-High, diagonal details).
The name of components refers to the combination of low-pass (L) and high-pass (H) filtering operations applied in both horizontal and vertical directions during 2D wavelet decomposition.

\begin{figure*}[htp!]
\centerline{\includegraphics[width=\linewidth]{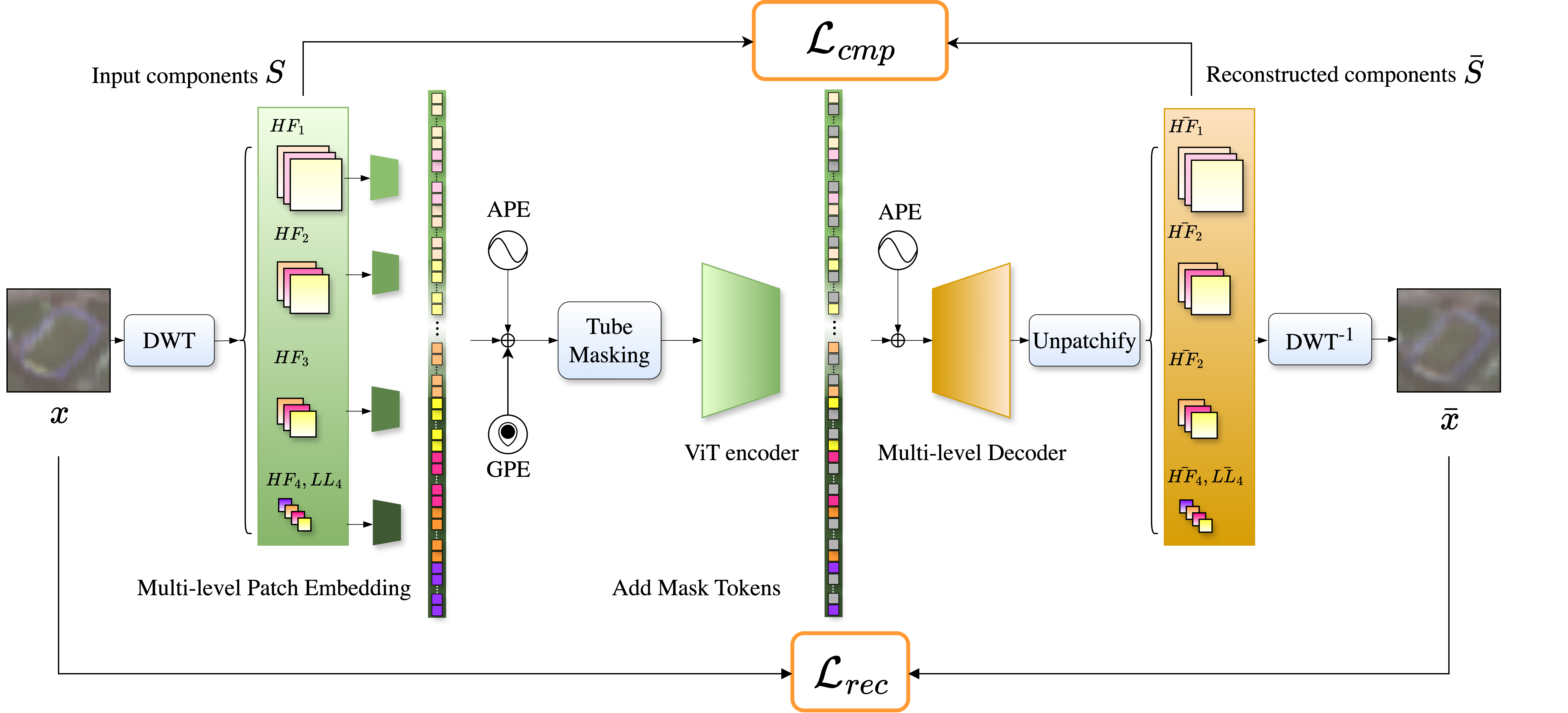}}
\caption{Illustration of WaveMAE architecture, the core design follows that of the MAE \cite{mae_paper}. We apply a 4-level DWT decomposition to the input image $x$ of resolution $(H,W)$ is obtaining 4 set of wavelet components at different scales of resolutions $(H/2, W/2), (H/4, W/4), (H/8, W/8)$ and $(H/16, W/16)$. 
Following this operation the components in $S$ are fed to the Multi-level patch embedding, here lighter colors stand for shallow decomposition levels and low-scale feature extraction, on the opposite darker colors mean deeper decomposition level and high-scale feature extraction.
After, the full sequence is fed to the MAE framework, the decoder reconstructs the masked wavelet components $\bar{S}$ at the same resolution specified previously. 
A smooth L1 loss is calculated between the input and the predicted wavelet components.
Additionally, the predicted components are used to calculate the inverse DWT, obtaining the reconstructed multispectral image $\bar{x}$.
An MSE loss is calculated between the latter and the input image $x$.
In both losses we attend only to the previously masked parts of the components/reconstructed image for the calculation.}
\label{image:architecture}
\end{figure*}

\subsection{Spherical Harmonics}
\label{subsection:sh}

Spherical harmonics (SH) describe functions on the surface of a sphere through orthogonal basis functions. SH are complex, continuous and limited functions defined as the angular part of the solution to Laplace's equation in spherical coordinates $Y_\ell^m(\theta, \phi)$, where $\ell$ is the degree ($\ell = 0, 1, 2, \ldots$), $m$ is the order ($m = -\ell, -\ell+1, \ldots, \ell-1, \ell$).

The equation can be explicited as:
\begin{equation}
Y_\ell^m(\theta, \phi) = \sqrt{\frac{2\ell+1}{4\pi}\frac{(\ell-|m|)!}{(\ell+|m|)!}} P_\ell^{m}(\cos\theta) e^{im\phi}
\end{equation}
where $P_\ell^{m}$ are the associated Legendre polynomials.
Intuitively, SH can be viewed as the spherical analogue of the Fourier series expansion.
For a periodic function defined on the circle (i.e., the interval $[0,2\pi]$), the standard approach is to decompose it into sines and cosines, which represent the natural vibration modes of the circle, each mode corresponding to an integer number of oscillations along the angular coordinate.
Analogously, for a function defined on a sphere $S^2$, we seek a decomposition into the natural vibration modes of the sphere. 
These modes are given by the spherical harmonics $Y_{\ell}^{m}(\theta,\phi)$, where the degree $\ell$ determines the overall number of oscillations on the sphere, while the order $m$ determines the azimuthal oscillation pattern.
Lower-degree harmonics ($\ell = 0, 1, 2$) represent smooth, global variations. Higher degrees capture increasingly fine-grained details and rapid variations on the sphere's surface. 
Any square-integrable function on the unit sphere can be expressed as follows:
\begin{equation}
f(\theta, \phi) = \sum_{\ell=0}^{\infty} \sum_{m=-\ell}^{\ell} c_\ell^m Y_\ell^m(\theta, \phi)
\end{equation}
with $(lat, lon) = (\theta, \phi) \text{ and } c_\ell^m$ is the weight associated to the orthogonal spherical harmonic basis functions.
The orthogonality property ensures that different basis functions are independent, enabling efficient computation and analysis.
Similarly to \cite{spherical_harmonics_paper}, only the real form of the spherical harmonics is selected for our application, and they are defined as:
\begin{equation}
Y_\ell^m(\theta, \phi) = 
    \begin{cases}
        (-1)^m \sqrt{2} \bar{P}^{m}_{\ell}(\cos{\theta})\sin{(|m|\phi)}) & \text{if } m < 0 \\
        \bar{P}^{m}_{\ell}(\cos{\theta}) & \text{if } m = 0 \\
        (-1)^m \sqrt{2} \bar{P}^{m}_{\ell}(\cos{\theta})\cos{(m\phi)}) & \text{if } m > 0
    \end{cases}
\end{equation}
where $\bar{P}^{m}_{\ell}(\cos{\theta})$ is the normalized associated Legendre polynomial.
In this way, the orthogonality and completeness properties are preserved while removing the complex term, making them more suitable for applications dealing with real-valued data on geographic coordinates, where $\theta$ represents latitude and $\phi$ represents longitude.
\section{Method}
\label{sec:method}

\subsection{WaveMAE}

The proposed method is tailored on the intuition of exploiting the properties of 2D Discrete Wavelet Transform, allowing our encoder to model multi-scale concepts in RS images and build richer features in different frequency resolutions.
Following this intuition, we built a Vision Foundation Model in an optical RS setup capable of exploiting high and low frequencies in multiple downstream tasks.
Additionally, in this section, we will introduce a novel Geo-conditioned Positional Encoding (GPE), tailored specifically for the RS scenario, which exploits Spherical Harmonics (SH) to encode geolocation coordinates as a positional embedding to shape the embedding space in a geographically-aware manner, with empirical improvements across several downstream tasks.

\subsection{Architecture}

\subsubsection{Pre-processing}
\label{section:pre-processing}

Figure \ref{image:architecture} shows the overall architecture of our method.
Firstly, the input multispectral images $x$ of spatial dimension  224$\times$224$\times C$ are used to perform a $N$ step DWT, with $C$ indicating the number of channels of the pretraining dataset.
Let $\mathrm{DWT}_N(x)$ denote the 2D discrete wavelet transform of an image $x$ up to $N$ levels.
At each level $j = 1,2,\dots,N$ we obtain high frequency wavelet components
$HL_j,\, LH_j,\, HH_j$ and, at the final level, $LL_N$.
Then, we define the set $S$ of all wavelet components as:
\begin{equation}
    \begin{aligned}
        S = &\bigcup_{j=1}^N S_j = S_1 \,\cup \, ... \, \cup\, S_N = \\
        &\{HL_1,\, LH_1,\, HH_1\} \,\cup \, ... \, \cup \,\{HL_N, LH_N, HH_N, LL_N\} 
    \end{aligned}
\end{equation}

We found $N=4$ to be the optimal solution, as demonstrated in our experiments, hence following the wavelet transform $\mathrm{DWT}_4(x)$ we obtain 13 components of which one is the Low-Low-frequency component (LL) and the remaining 12 are the high-frequency components at different scales, 3 per each level of decomposition.
More intuitively, the Low-Low component $LL_4$ is obtained by applying a $16\times$ downscaling to the original image $I$. For each decomposition level $i \in \{1, 2, 3, 4\}$, the high-frequency components $\{HL_i, LH_i, HH_i\}$ are obtained at downscale factor $2^i$:
\begin{align*}
HL_1, LH_1, HH_1 &\rightarrow 2\times \text{ downscale} \\
HL_2, LH_2, HH_2 &\rightarrow 4\times \text{ downscale} \\
HL_3, LH_3, HH_3 &\rightarrow 8\times \text{ downscale} \\
HL_4, LH_4, HH_4 &\rightarrow 16\times \text{ downscale}
\end{align*}
This mechanism allows retaining frequencies at different scales, thus having access to both fine-grained details and insights of larger parts of the images.
More precisely, with the considered image input resolution, the set of input components $S$ will have different spatial resolutions: $S_1 = 112 \times 112$, $S_2 = 56 \times 56$, $S_3 = 28 \times 28$ and $S_4 = 14 \times 14$.

\subsubsection{Multi-level Patch Embedding}
\label{section:multi_level_patch_embedding}

\begin{figure*}[tbh]
\centerline{\includegraphics[width=\linewidth]{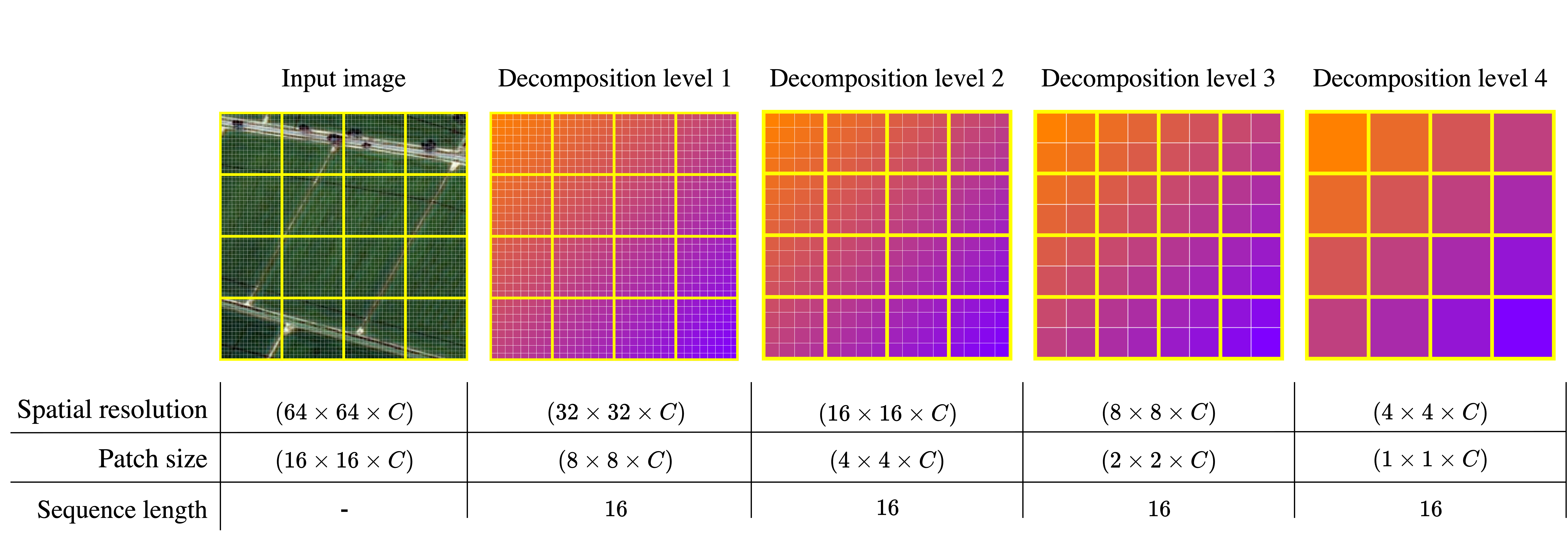}}
\caption{In this figure we show the mechanism of our Multi-level Patch Embedding. 
Considering an input image of spatial dimension $64\times 64$, for $N=4$ decomposition level we obtain the set of DWT components of resolution specified in the table.
Scaling the patch size accordingly to the component's spatial resolution, allows us to attend to the same information of the input image in the correspondent spatial location at different decomposition levels.
White color is used to show the grid of pixels while yellow outlines the patches borders.
}
\label{image:multi_level_patch_embedding}
\end{figure*}

The core design principle of WaveMAE is to encourage the encoder to learn the relationship between tokens which belong to the same spatial location at multiple decomposition levels, in order to enable an association between frequency components at different scales.
To accomplish this, we designed our Multi-level Patch Embedding as a set of convolutions, one per decomposition level and each independent from the others, with patch size scaled down according to the downscale factor of the level component. The process can be understood more intuitively in Fig. \ref{image:multi_level_patch_embedding}.
By setting the maximum number of decomposition to $N=4$, the patch size of each level component starting from shallow ($S_1$) to deepest ($S_4$) is 8$\times$8, 4$\times$4, 2$\times$2 and 1$\times$1, making the sequence length equal for each component at every decomposition level.
This patch size defines the kernel dimension of the patch embedding layer at that level.
This approach guarantees that the amount of input information contained in each token of a specific spatial location attends to the pixel values of a patch size of 16$\times$16 in the original image $x$ in that same spatial location.
For example, as illustrated in Fig. \ref{image:multi_level_patch_embedding}, the top-left patch of each component across decomposition levels (columns 2 to 5) encodes wavelet information at progressively different scales of the same top-left region in the original image (first column).

\subsubsection{Positional Encoding}
\label{subsection:pe}

Before passing the sequence of tokens through the WaveMAE encoder, we proceed to add positional encodings to the tokens. 
To discriminate the position of the tokens in the sequence, we used the sinusoidal Absolute Positional Encoding (APE) due to its proven effectiveness \cite{attention_paper}.
\begin{figure}
\centerline{\includegraphics[width=\linewidth]{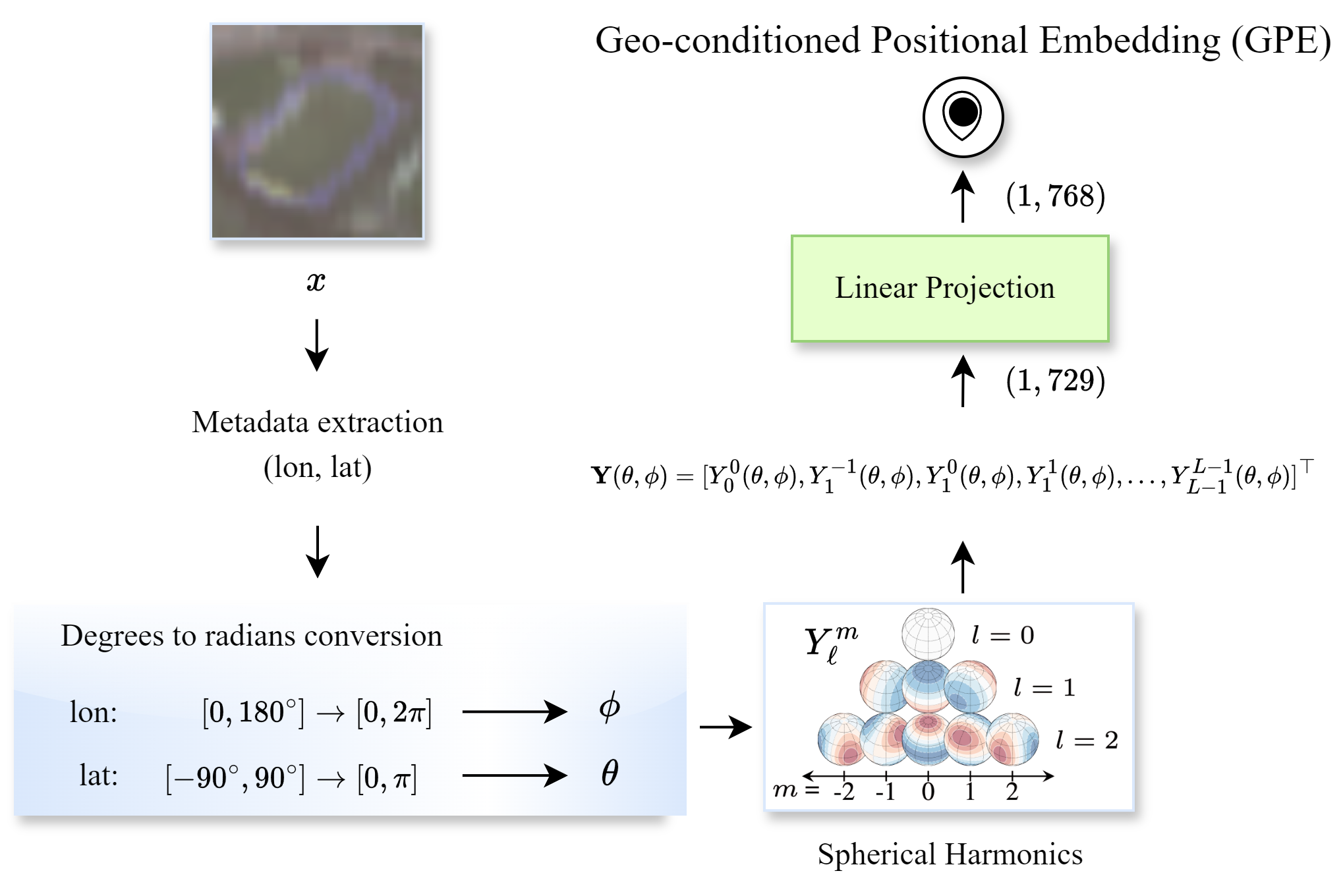}}
\caption{The illustration shows how we obtain the novel Geo-conditioned Positional Encoding. Firstly, from any image $x$ the metadata associated are read to extract the geospatial coordinates in terms of latitude and longitude. Since coordinates are stored in degree we apply a degree to radians conversion and calculate a set of Spherical Harmonics as specified in Section \ref{subsection:pe}. After this step we apply a linear projection to bring the encoding to the embedding dimension $D=768$ of our encoder.}
\label{image:gpe}
\end{figure}
Additionally, in order to incorporate geographical information into the embedding space, we design a novel Geo-Conditioned Positional Encoding (GPE) that leverages Spherical Harmonics to encode the geolocation of each input image, as shown in Fig. \ref{image:gpe}. 
This approach allows patches that belong to similar geographical regions to be positioned closer together in the embedding space.
For any pretraining image $x$, we use associated metadata containing latitude and longitude coordinates $(\text{lat}, \text{lon})$ and, as a first step, we convert these geographical coordinates from degrees to radians:

\begin{equation}
    \begin{aligned}
        \theta &= \text{deg2rad}(\text{lat} + 90\text{\textdegree}) \in [0, \pi]\\
        \phi &= \text{deg2rad}(\text{lon} + 180\text{\textdegree}) \in [0, 2\pi] \\
    \end{aligned}
\end{equation}

\noindent where $\theta$ is the polar angle and $\phi$ represents the azimuth in the coordinate system.
For each degree $\ell=0,\ldots,L-1$ and order $m=-\ell,\ldots,\ell$, we evaluate the spherical harmonic $Y_\ell^{m}(\theta,\phi)$.
Increasing the cutoff $L$ appends higher-degree components (larger $\ell$ and $|m|$), which have shorter angular wavelengths and therefore increase the resolvable spatial detail, low-degree harmonics capture broad geographic trends, while higher-degree harmonics encode fine, local variations \cite{spherical_harmonics_paper}.
We set $L = 27$ to ensure a favorable trade-off between harmonic resolution and computational efficiency.
This parameterization yields a total of:
$\sum_{\ell=0}^{L-1} (2\ell + 1) = L^2 = 729 \text{ harmonic coefficients.}$
The complete spherical harmonics representation is constructed by concatenating the harmonics calculated previously in the following way:
\begin{equation}
    \begin{aligned}
        \mathbf{Y}(\theta, \phi) = [ Y_0^0(\theta, \phi), Y_1^{-1}(\theta, \phi), Y_1^0(\theta, \phi), Y_1^1(\theta, \phi), ..., \\
            Y_{L-1}^{L-1}(\theta, \phi)]^\top \in \mathbb{R}^{L^2}
    \end{aligned}
\end{equation}
Finally, the GPE is obtained by projecting this high-dimensional spherical harmonics representation to the transformer's embedding dimension $D = 768$ through a linear projection.
The resulting GPE is added to all tokens in the sequence, ensuring that the spatial embedding encodes both local patch position (via APE) and global geographical context (via GPE).
This dual encoding enables the model to learn representations that are aware of both fine-grained spatial relationships within images and broader geographical patterns across the globe.

\subsubsection{Masking Strategy}
\begin{figure}
\centerline{\includegraphics[width=\linewidth]{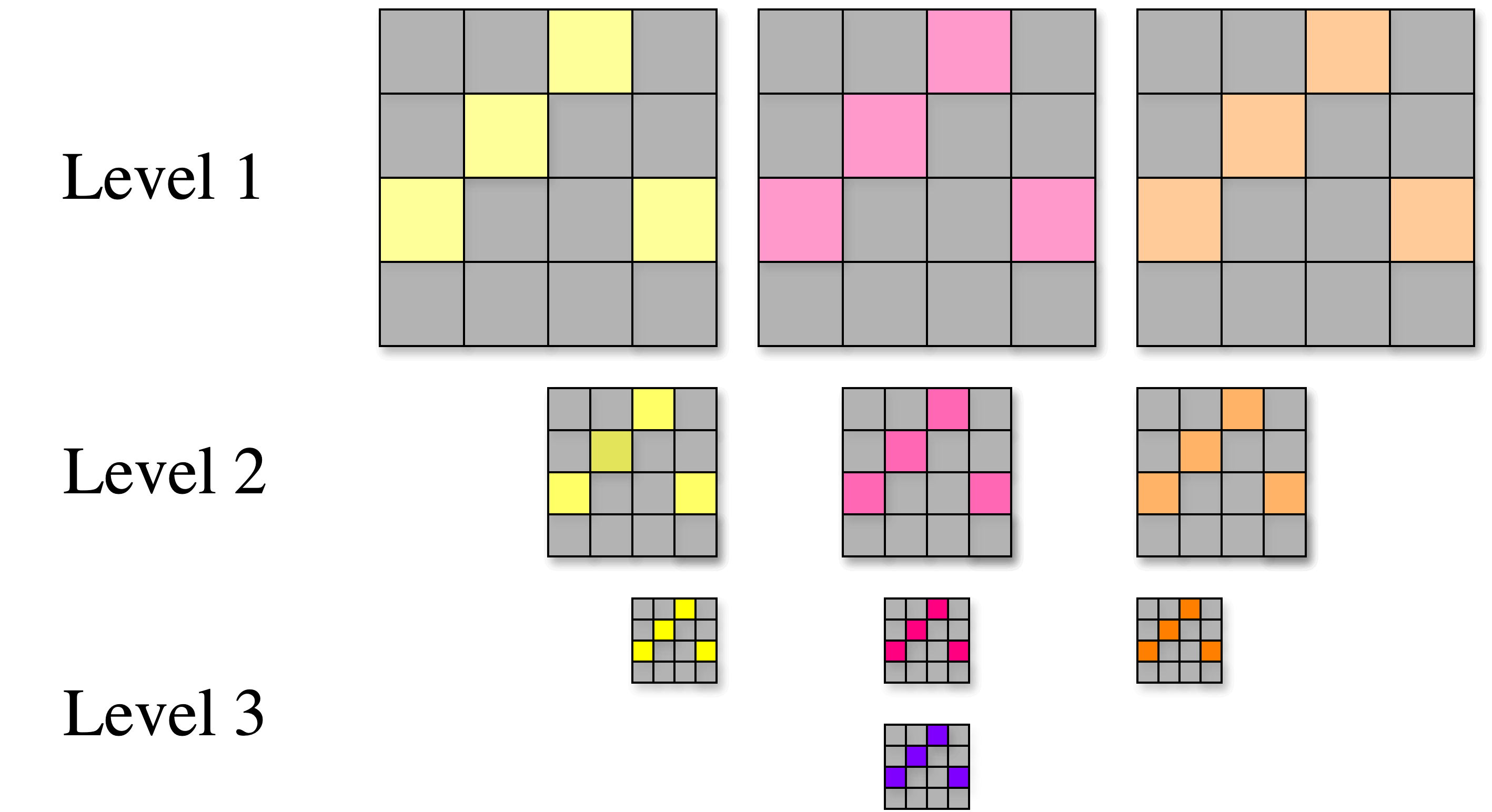}}
\caption{Tube Masking strategy: supposing a 3 level DWT decomposition each component in all levels are masked spatially in the same way. From top to bottom, in each row, we have shallow to deeper level of decomposition, components are indicated in different colors [\textcolor{ll_color}{LL}, \textcolor{hl_color}{HL}, \textcolor{lh_color}{LH}, \textcolor{hh_color}{HH}], gray patches are masked.}
\label{image:tube_masking}
\end{figure}
The masking strategy adopted exploits the spatial alignment of the tokens across all the components and decomposition levels generated by our Multi-level Patch Embedding, thus allowing the adoption of a random Tube Masking protocol \cite{videomae_paper}.
This type of masking allows to relax the spatial redundancy while preserving the frequency and decomposition correlation by masking all the wavelet components in the same way as shown in Fig. \ref{image:tube_masking}.
The mask ratio is set at 75\% of the total sequence, aligning with other self-supervised methods \cite{satmae_paper}\cite{satmae++_paper}.\\
More formally, we consider the set T of patch tokens $\mathcal{T}=\{\, t_{i,j}^{\,l} : i=1,\dots,n_H,\; j=1,\dots,n_W,\; l=1,\dots,L \,\}$, where $(i,j)$ represents spatial location, $l$ the decomposition level, $n_H$ and $n_W$ the number of tokens, respectively, on the height and width of the original image $x$, while $L$ is the maximum decomposition level taken into consideration.
We draw a random spatial mask 
\begin{center}
    $M \subseteq \{1,\dots,H\}\times\{1,\dots,W\}$ with $|M|=\rho HW$ and $\rho=0.75$ (masking ratio). 
\end{center}
The retained tokens are:
\begin{equation}
    T=\bigcup_{(i,j)\notin M}\{\, t_{i,j}^{\,l} : l=1,\dots,L \,\}
\end{equation}
Our intuition behind this design choice, in contrast to a completely Random Masking, was related to the convergence speed.
In our preliminary experiments, adopting the Random Masking did not yield satisfactory reconstruction results, on the opposite, Tube Masking led the model to learn the relationship between patches in the same spatial location faster, thereby stabilizing the training.
We hypothesize that, for each unmasked token in a specific spatial location, the model can attend to the tokens belonging to any decomposition level and any wavelet component in that same spatial location, allowing access to the full hierarchical structure of wavelet decomposition, thus fastening the convergence.

\subsubsection{Multi-level Reconstruction}

\begin{figure*}
    \centerline{
        \includegraphics[width=\linewidth]{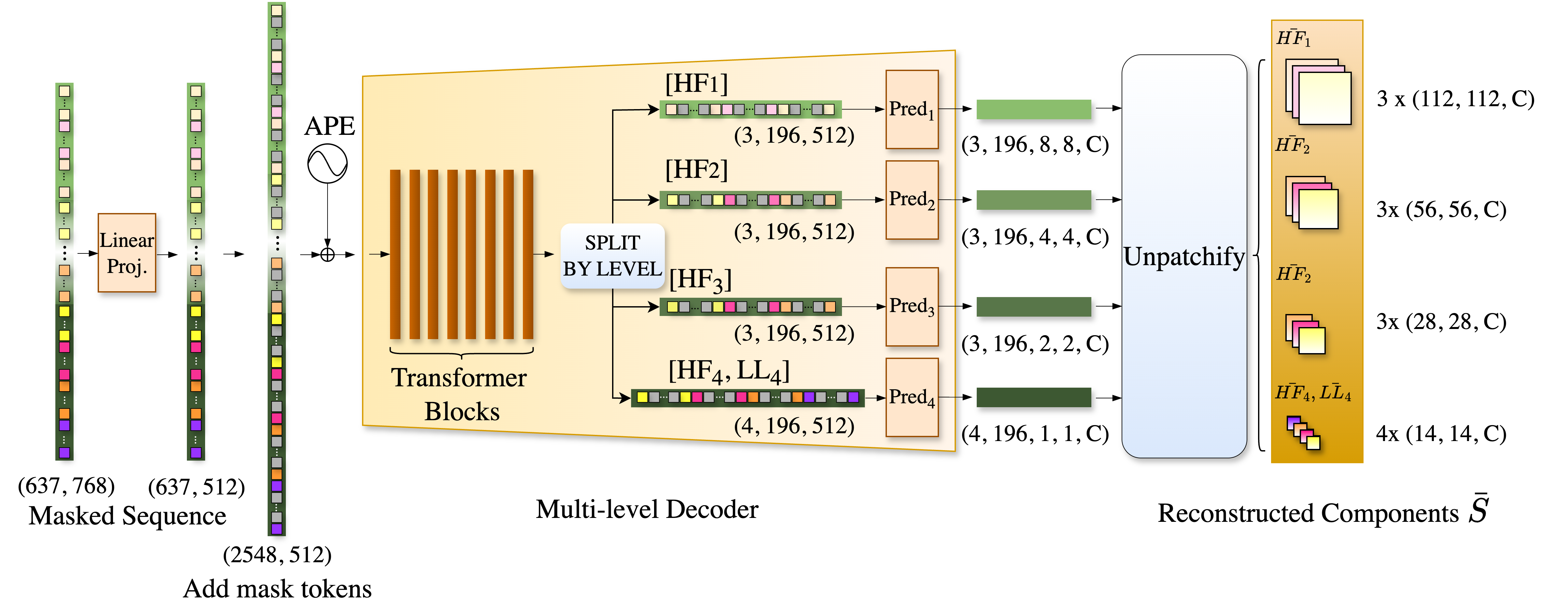}
    }
    \caption{
        In this illustration, we present our Multi-level Decoder. Decoding begins by projecting the masked sequence embedding to the decoder’s embedding dimension $\hat{D} = 512$. A set of learnable mask tokens is then inserted in place of the previously masked tokens. After adding absolute positional embeddings (APE), the full sequence is processed by $8$ Transformer blocks.  
        To predict the pixel values of the components at different levels, we employ four independent Pred modules, one for each level.
        For this purpose, the full sequence is split into level-specific subsequences, each containing all tokens of the corresponding level. 
        Owing to the design of our Multi-level Patch Embedding, each component has a fixed sequence length of $196$ tokens at any level. 
        Consequently, the $4^th$ (deepest) level sequence consists of $784$ tokens, as it includes both $HF_4 = \{HL_4, LH_4, HH_4\}$ and $LL_4$, while the other levels contain $588$ tokens each, corresponding only to the high-frequency components. 
        It is worth noting that the Pred module is composed by 3 separate MLPs (4 in the deepest level), where each of them processes only the tokens of one component.
        Each Pred module maps the embedding dimension to the patch size, $512 \rightarrow p_H \times p_W$, where $p_H$ and $p_W$ denote the patch height and width, respectively. 
        These values depend on the decomposition level, as described in Section~\ref{section:multi_level_patch_embedding}. Finally, an unpatchify operation is applied to obtain the set of reconstructed components $\bar{S}$.
    }
    \label{image:multi_level_decoder}
\end{figure*}

Subsequently to the encoding phase, we linearly project the embedding to decoder hidden dimension $\hat{D}=512$ and we add back to the sequence a set of learnable mask tokens in place of the previously masked tokens, following the regular MAE workflow \cite{mae_paper}. After that, we sum the APE to the full sequence to distinguish the position of the single tokens inside the sequence.
The Geo-conditioned Positional Encoding (GPE) is injected only into the encoder inputs and is not appended to the decoder sequence. This prevents the decoder from accessing explicit geolocation information during reconstruction, potentially facilitating the task.

In Fig.~\ref{image:multi_level_decoder}, we show WaveMAE's decoder structure.
Unlike the original MAE \cite{mae_paper}, this decoder does not directly reconstruct the image $\bar{x}$ in the pixel domain, instead, it aims to reconstruct a set of wavelet components named $\bar{S}$, which are then used to derive the reconstructed image $\bar{x}$ through the inverse transform:
\begin{equation}
    \bar{x} = DWT_{N}^{-1}(\bar{S})
\end{equation}
To this end, the linear projection following the decoding phase is composed of $L$ independent Pred modules, mirroring our Multi-level Patch Embedding.
Each Pred module at the end of the Multi-level Decoder will reconstruct all the wavelet components in that level.
By concatenating all the wavelet components we obtain, in the end, $\bar{S}$.

\subsubsection{Losses}

Though intuitively minimizing the reconstruction of the sole components in $\bar{S}$ would be enough to guarantee a good quality of the reconstructed target, we found that this is not the case.
For this reason, we adopted two losses to achieve better reconstruction details:
\begin{equation}
    \mathcal{L}_{tot} = \mathcal{L}_{rec} + \mathcal{L}_{cmp}
\end{equation}
In both losses we attend to the mask $M$ used for each image, meaning that only previously masked spatial locations are used for the calculation.
$\mathcal{L}_{rec}$ minimizes the reconstruction error between the original image $x$ and the inverse DWT-transformed image using the predicted components $\bar{x} = DWT_4^{-1}(\bar{S})$, being $\bar{S}$ the set of predicted wavelet components:
\begin{equation}
    \mathcal{L}_{\text{rec}} \;=\; \frac{1}{P} \sum_{i=1}^{P} \left( x_i - \bar{x}_i \right)^2
\end{equation}
\noindent where $P$ is the number of pixels masked before the encoder. 
This loss is of paramount importance as it induces the encoder to produce a representation that contains information fit for the inverse DWT, pushing high frequencies even in the shallow levels of decomposition.
On the other side, we employ $\mathcal{L}_{cmp}$ as a regularization loss using the predicted components and calculating a smooth L1 loss: 

\begin{equation}
    \mathcal{L}_{\text{cmp}} \;=\; \frac{1}{|M|} \sum_{p \in M} 
    \mathrm{SmoothL1}\!\big(S_p, \bar{S}_p\big)
\end{equation}
This loss empirically shows benefits on several downstream tasks when used in conjunction with our GPE.

\section{Experiments}
\label{sec:experiments}

To ensure a fair evaluation, we carefully isolated the sources of performance improvements from factors such as the type and quantity of pretraining data, thereby allowing us to directly address the efficacy of the pretraining strategy alone.
For this reason, we adopted a common pretraining dataset and retrained all methods under consideration on this shared resource, establishing a consistent and unbiased setup for comparison. 
To rigorously validate the effectiveness of the pretraining techniques, we deliberately excluded suboptimal downstream datasets often employed in prior works and instead relied on the PANGAEA benchmark \cite{pangaea_paper}, which provides a diverse and representative collection of tasks. 
In total, we selected five datasets spanning four downstream objectives: semantic segmentation, change detection, regression, and multi-label classification. 
Further details on each dataset are reported in Section~\ref{subsection:dataset}.

All methods considered in our study belong to the family of masked image modeling for remote sensing imagery, as these approaches are known to better exploit high-frequency information for representation learning \cite{park2023ssl}. Such frequencies are both challenging and crucial to capture for fine-grained analysis of remote sensing data, given its inherently large spatial scale.
Within this context, we aim to show that optical encoding particularly benefits from the introduction of wavelet decomposition, which provides a structured and scale-aware representation of high-frequency content that is often overlooked by conventional approaches.
With this in mind, we restricted our comparison to foundation models that adopt a masked autoencoding paradigm and can be applied exclusively to optical remote sensing data.
This choice allows us to directly assess the impact of architectural design choices and pretraining strategies within a homogeneous setting, without confounding contributions from additional modalities such as Synthetic Aperture Radar (SAR).
Accordingly, we include in our comparison representative approaches such as MAE \cite{mae_paper}, which serves as the original formulation of masked autoencoders for natural images and provides a strong baseline for transfer to other domains.  
On top of this, SatMAE \cite{satmae_paper} adapts the MAE framework to remote sensing by incorporating multispectral data and exploiting the inherent redundancy across spectral bands, while SatMAE++ \cite{satmae++_paper} further improves the reconstruction objective through additional multi-scale constraints, yielding richer representations for downstream tasks.
In order to expand the assessment, we also consider SpectralGPT \cite{spectralgpt_paper}, a recent large-scale transformer designed specifically for multispectral imagery, which introduces spectral priors to better model fine-grained inter-band relationships.
Together, these baselines include both general-purpose and domain-specific instantiations of masked autoencoders, thereby providing a comprehensive and balanced point of reference against which to evaluate the contributions of our WaveMAE.

\subsection{Datasets}
\label{subsection:dataset}
Our experimental setup focuses on optical data, and in particular on multispectral imagery, owing to its proven effectiveness and the additional information it provides beyond standard RGB bands. For this reason, we adopt fMoW-S2 \cite{fmow_paper} as the baseline pretraining dataset for all models evaluated in this section.

\textbf{fMoW-S2} (Functional Map of the World – Sentinel-2) is the Sentinel-2 variant of the IARPA fMoW collection.
It contains 882,779 images covering hundreds of countries and collects the full set of Sentinel-2 spectral bands (commonly reported as B1--B12 plus B8A, i.e. 13 spectral bands) over 62 distinct categories. 
These include four 10\,m bands (B2, B3, B4, B8), six 20\,m bands (B5, B6, B7, B8A, B11, B12) and three 60\,m bands (B1, B9, B10).
To rigorously assess transfer learning capabilities, we follow the PANGAEA benchmark protocol \cite{pangaea_paper} and finetune all pretrained models on five downstream domains spanning four tasks. PANGAEA provides a standardized evaluation protocol that covers diverse datasets, tasks, sensors, and resolutions. \\
Specifically, we finetune on the following datasets:

\textbf{MADOS.} MADOS \cite{mados_paper} is a Sentinel-2–based semantic segmentation dataset for marine pollution, focusing on oil spills and marine debris. The dataset is assembled from Sentinel-2 and therefore provides the Sentinel-2 multispectral complement (i.e. B1, B2, B3, B4, B5, B6, B7, B8, B8A, B11, B12, i.e., 11 bands in total). MADOS leverages the multispectral information to discriminate pollution classes (e.g., Oil, MarineDebris) and is well suited to evaluate spectral sensitivity and fine-grained segmentation performance under marine conditions.

\textbf{HLSBurnScars.} HLSBurnScars \cite{hlsburnscars_paper} contains burn-scar segmentation scenes derived from Harmonized Landsat–Sentinel (HLS) products. The dataset provides harmonized optical reflectances combining Landsat and Sentinel-2 inputs and commonly publishes a subset of bands (i.e., B2, B3, B4, B8A, B11, B12, 6 bands in total) at the HLS harmonized resolutions. Scenes are typically provided at $512\times512$ pixels and the public release contains $804$ scenes across the contiguous United States (2018–2021), making it a useful benchmark to test robustness across acquisition conditions and harmonized cross-sensor inputs.

\textbf{SpaceNet7.} SpaceNet7 \cite{spacenet_paper}, also referred to as the Multi-temporal Urban Development SpaceNet (MUDS) challenge, addresses multi-temporal urban change detection with per area-of-interest (AOI) time-series stacks and building footprint labels. For the MUDS challenge, the imagery is provided by PlanetScope (Planet), which typically has a ground sampling distance of approximately $3\text{--}4\ \text{m}$ and supplies regular RGB bands (i.e., B2, B3, B4). The dataset structure (monthly stacks and standardized patches per AOI) enables evaluation of temporal change detection and building-level localization under medium resolution.

\textbf{BioMassters.} BioMassters \cite{biomassters_paper} is a regression benchmark for above-ground biomass estimation that combines Sentinel-1 Synthetic Aperture Radar (SAR) data and Sentinel-2 Multispectral Imagery (MSI) time series with LiDAR-derived reference labels. Sentinel-1 contributes SAR channels (i.e., VV, VH) while Sentinel-2 provides the full optical band set (i.e., B2, B3, B4, B5, B6, B7, B8, B8a, B11, B12, 12 bands in total). 
The dataset comprises 310,000 patches (each patch covering roughly a $2{,}560\times2{,}560\ \text{m}$ area) with per-patch biomass targets in Finland, offering a multi-modal testbed for time-series regression.

\textbf{BigEarthNet.} BigEarthNet \cite{ben_paper} is a large-scale multi-label scene classification benchmark based on Sentinel-2 Level-2A products. 
The dataset supplies atmospherically corrected Sentinel-2 patches and therefore includes the standard MSI bands (B1–B12, B8A; 13 bands). 
BigEarthNet v2 contains 549,488 patches (commonly used as a large-scale training and evaluation corpus).
Typical preprocessing maps the 10\,m bands to patches of about $120\times120$ pixels (with corresponding sizes for 20\,m and 60\,m bands), and annotations are provided as multi-label categories (19 classes).

Together, these datasets span multiple spatial resolutions, spectral bands, and downstream objectives (segmentation, change detection, regression, and multi-label classification), providing a comprehensive and balanced suite for evaluating the transferability of pretrained remote sensing representations.
Since our focus is exclusively on the optical scenario, we discard Sentinel-1 image patches and retain only Sentinel-2 multispectral data when available, or RGB imagery otherwise.

\subsection{Pretraining Details}

As anticipated, we adopt fMoW-S2 \cite{fmow_paper} as the pretraining dataset for all methods considered in our comparison. 
Following prior works \cite{spectralgpt_paper}, the preprocessing pipeline applied to the multispectral images consists of normalization to a standardized range in $[0,1]$, followed by a random crop with scale factor uniformly sampled between $0.2\times$ and $1.0\times$. 
Subsequently, each crop is resized to a fixed spatial resolution $I=(I_h,I_w)=(224,224)$, and a random horizontal flip with probability $p=0.5$ is applied, following established protocols \cite{satmae_paper, spectralgpt_paper}.
The number of input spectral bands is set to $C=13$ for all methods, comprising all spectral bands of Sentinel-2 data, with the exception of SpectralGPT~\cite{spectralgpt_paper} where, due to the patch embedding design proposed in the original paper, we set $C=12$ excluding band B1.

All models are pretrained for 50 epochs using the AdamW optimizer with a learning rate of $10^{-4}$. 
For our WaveMAE, an additional preprocessing step is performed: a Discrete Wavelet Transform (DWT) with decomposition level $N=4$ is applied, yielding 13 wavelet components that are stacked to form the input sequence. 
We employ the Haar wavelet basis for all decompositions.
As a baselines, the patch size is set to $P=(p_h,p_w)=(16,16)$, which corresponds to $(I_h/p_h)\times(I_w/p_w)=196$ spatial tokens for all methods considered for comparison. 
In contrast, WaveMAE adopts a multi-level patch embedding (Section~\ref{sec:method}), where input components at each decomposition level are spatially downsampled by a factor of $2\times$ with respect to the original image resolution.
To preserve token alignment across decomposition levels, we scale accordingly the base patch size to $\hat{P}=(\hat{p_h},\hat{p_w})=(8,8)$, which is equivalent to a $(16,16)$ patch in the original image space (as previously illustrated in Fig.~\ref{image:multi_level_patch_embedding}).
This allows to have in the first level $(\hat{I_h},\hat{I_w})=(112,112)$, thus $(\hat{I_h}/\hat{p_h})\times(\hat{I_w}/\hat{p_w})=196$ spatial tokens, which is consistent with the baseline.
The number of spatial tokens remains constant across subsequent decomposition levels, since both the component resolution and the patch size are downscaled by a factor of $2\times$ at each level.

Geographic coordinates for WaveMAE’s Geo-conditioned Positional Encoding are extracted directly from the GeoTIFF metadata provided in fMoW-S2 and processed as described in Section~\ref{sec:method}. 
Unless differently specified, masking ratio is set to 75\% by default for all methods, aligning with prior works \cite{mae_paper}.
All pretraining experiments were conducted on 2 NVIDIA L40S GPUs, ensuring consistency across all experiments.  

\subsection{Downstream Tasks}

To rigorously assess the transfer learning capabilities of the pretrained encoders, we evaluate all methods on a diverse set of downstream tasks drawn from the PANGAEA benchmark \cite{pangaea_paper}. 
In this stage, only the encoder weights obtained during pretraining are transferred, while the decoder is discarded. For each downstream task, a task-specific head is trained on top of the pretrained encoder, with hyperparameters meticulously aligned across methods to ensure experimental fairness and eliminate confounding factors arising from optimization differences.

For all tasks except classification, feature maps from encoder layers ${3,5,7,11}$ are aggregated to form a feature map enriched with low, intermediate, and high-level features as input to the decoder, following established practices in dense prediction tasks \cite{densepredictionvit_paper}.
In all methods except WaveMAE and MAE, each output layer produces features of dimension $(B, N, L, D)$ which are aggregated by summation over the $N$ dimension, representing the number of spectral band groups utilized by the respective method \cite{satmae_paper, satmae++_paper, spectralgpt_paper}, followed by layer normalization applied along the channel dimension $D$. 
This aggregation yields feature representations of dimension $(B, L, D)$, which undergo a permutation over the spatial and channel dimensions before being reshaped to form the final 2D feature map $(B, D, f_h, f_w)$, where $f_h = I_h/p_h$ and $f_w = I_w/p_w$ correspond to the spatial dimensions derived from the input image resolution $(I_h, I_w)$ and patch size $(p_h, p_w)$.

In WaveMAE, the feature extraction process requires specialized handling due to the multi-component wavelet representation. At each output layer, representations maintain dimension $(B, N, L, D)$ where $N$ corresponds to the number of wavelet components. 
The low-frequency LL component is extracted separately from the high-frequency (HF) components, with independent layer normalization applied to each subset to account for their distinct statistical distributions and prevent the suppression of low-frequency information through inappropriate feature mixing. 
The HF components are subsequently aggregated via summation over the $N$ dimension and layer normalized independently. 
The processed low-frequency and high-frequency representations are then concatenated, yielding features of dimension $(B, 2, L, D)$, which undergo final aggregation through summation over the component dimension followed by layer normalization to produce the standard $(B, L, D)$ representation suitable for 2D feature map reconstruction.
Moreover, GPE was applied only to datasets providing geolocation metadata. Consequently, MADOS and BioMassters did not yield geolocation information during finetuning, and the effect of GPE was confined to the pretraining stage.
Regular MAE maintains dimension $N=1$ by default, requiring no additional preprocessing operations before feeding the features to the downstream decoding head. 
All downstream experiments are conducted on a single NVIDIA L40S GPU to ensure consistent computational conditions across evaluations.

\textbf{Semantic Segmentation}. For semantic segmentation, we adopt the UperNet \cite{upernet_paper} architecture following the protocol defined in the PANGAEA benchmark. 
Performance is measured in terms of mean Intersection over Union (mIoU), which is the standard metric for semantic segmentation as it accounts for both precision and recall across spatially distributed classes. 
The training setup uses a cross-entropy loss and the AdamW optimizer with learning rate $1\times 10^{-4}$, betas $(0.9,0.999)$, weight decay $0.05$, and a batch size of 32. 
Training is performed for 80 epochs with validation every 20 epochs, selecting the best checkpoint based on the evaluation metric performance on the validation set. 
For images larger than the model input size, sliding-window inference is applied. 
We employ a multi-step learning rate scheduler with milestones at $[0.6,0.9]$ of the total training epochs. 
This setup is applied to both MADOS and HLSBurnScars datasets.

\textbf{Regression}. For regression tasks, we employ a Regression UperNet head, reusing the same optimizer, scheduler, and intermediate layer aggregation as in segmentation. 
The training is conducted for 50 epochs with validation occurring every 10 epochs, using a batch size of 32 and an MSE loss. 
The evaluation metric is the root mean squared error (RMSE), computed on the test set using the best checkpoint selected via validation. 
RMSE is adopted as it directly penalizes large deviations between predicted and reference values, which is crucial for accurately estimating continuous biophysical variables such as above-ground biomass (AGBM). 
This setup is applied to the BioMassters dataset.

\textbf{Multilabel scene classification}. 
For multilabel scene classification, we discard the 2D decoder structure and instead train a lightweight classification MLP on top of the highest-level encoder features. 
The classification head is composed of an initial \texttt{Linear} layer, followed by a \texttt{BatchNorm1d} normalization, a non-linear \texttt{ReLU} activation, and a final \texttt{Linear} layer that outputs the class logits.
Training is performed for 20 epochs with validation every 5 epochs, using a batch size of 256 and a Binary Cross Entropy loss. 
This task uses the same optimizer and scheduler as the previous.
The evaluation metric is mean average precision (mAP), which provides a robust measure of performance in the multilabel setting by accounting for both class imbalance and partial label correlations. 
This setup is applied to the BigEarthNet dataset.

Across all downstream tasks, we ensure consistent evaluation by standardizing the training protocols, optimizer settings, and feature usage across methods, while varying only the task-specific heads as required by the problem. 
This strategy allows us to fairly isolate the impact of the pretrained representations obtained by MAE, SatMAE, SatMAE++, SpectralGPT, and our proposed WaveMAE.

\subsection{Ablation Study}
\label{subsec:ablation_study}

To evaluate the individual contributions of each architectural component, we conduct a complete ablation study that isolates the impact of every major element on the final model performance. 
This analysis provides insights into the effectiveness of our design decisions and establishes the necessity of each proposed component within the overall framework.
A single pretrained model for each configuration is evaluated across all five downstream datasets described in Section \ref{subsection:dataset} for each ablation experiment. 
We report the average performance across five independent experiments conducted with different random seeds for each model configuration. 
This methodology ensures that our conclusions are not influenced by initialization conditions or training instabilities.
In these ablation experiments, we establish a controlled experimental environment that prioritizes computational efficiency during the extensive ablation phase while maintaining sufficient representational capacity to demonstrate the relative contributions of each component. For this reason, except for the token size analysis in Section \ref{subsec:token_size}, we adopt a base patch size of $(p_h,p_w)=(16,16)$ for the first decomposition level, scaled down along component resolution according to the decomposition level, following Section \ref{section:multi_level_patch_embedding}. 
Given the multi-level structure of our wavelet-based approach, this configuration results in a sequence length that is shorter than the one employed in our final model architecture. 
Specifically, for input images of resolution $224\times224$, the wavelet components are spatially downsampled to $112\times112$ resolution at the first level. 
With a patch size of $16\times16$, this yields $\frac{112}{16} \times \frac{112}{16} = 7 \times 7 = 49$ spatial tokens per decomposition level, compared to the $196$ tokens used in our final configuration. 

Our ablation study follows an incremental approach. 
We structure the analysis to address five key research questions in sequential order regarding the fundamental efficacy of the Discrete Wavelet Transform (DWT) by varying the decomposition level, the effect of our novel Geographic Positional Encoding (GPE), the impact of masking ratio on pretraining effectiveness, the transfer learning gains provided by our self-supervised approach and, finally, the influence of token size on model performance.

\subsubsection{Decomposition level}

\begin{figure}
\centerline{\includegraphics[width=\linewidth]{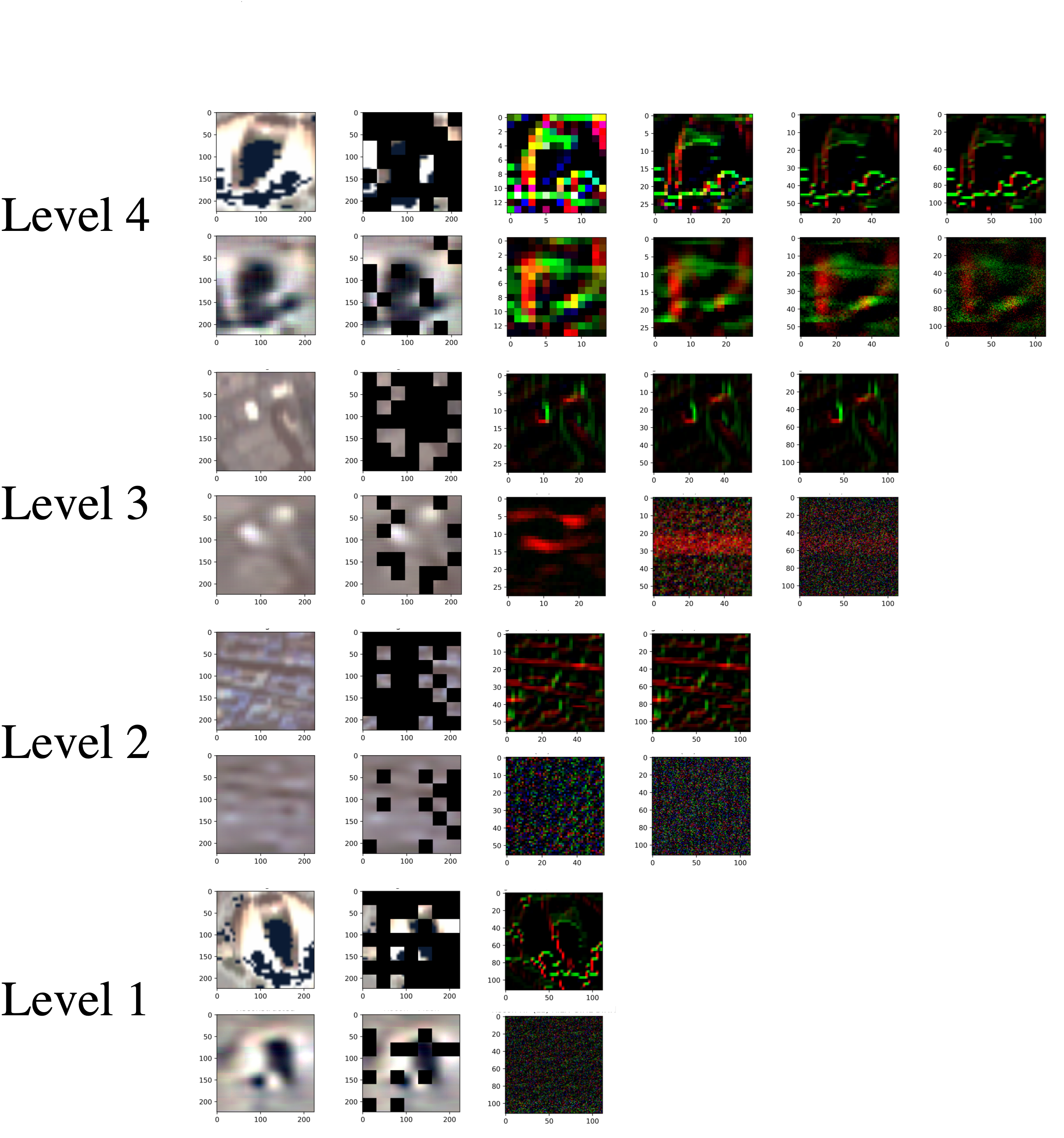}}
\caption{In this Figure we show qualitative reconstructions by considering four decomposition levels pretraining configurations at epoch 50. 
For each level, the \emph{top} row displays (left-to-right) the target image, the unmasked portion of the target, and the target high-frequency component for the band corresponding to the red channel (fMoW-S2 band 2). 
To distinguish between high frequencies components we show in red LH component (vertical details), green for HL (horizontal) and blue for HH (diagonal details).
The \emph{bottom} row reports the corresponding outputs produced by the model: the reconstructed image, the previously masked region reconstructed by the model, and the reconstructed high-frequency component for the same spectral band. 
Across levels 1–3 the encoder struggles to recover fine high-frequency details and instead relies on low-frequencies cues to approximate the target image. 
At level 4, conversely, the lower-resolution high-frequency components are easier to learn, and these coarse high-frequency cues support improved reconstruction of inner-level components. 
Overall, the figure illustrates the progressive difficulty of recovering high-frequency information at finer scales and highlights the role of coarse-level HF components in stabilizing reconstruction.
}
\label{image:level_reconstruction_quality}
\end{figure}

To establish the optimal depth for wavelet decomposition in our WaveMAE framework, we systematically evaluate the effect of varying decomposition levels on downstream task performance across our comprehensive benchmark suite. 
This analysis is particularly crucial, as the decomposition level directly controls the hierarchical structure of the wavelet representation, determining both the granularity of frequency separation and the complexity of the reconstruction objective during pretraining.
The reconstruction difficulty inherently increases with lower decomposition levels, as fewer frequency components are available to capture the full spectral and spatial complexity of multispectral remote sensing imagery. 
Conversely, higher decomposition levels provide a richer hierarchical structure that enables the model to learn multi-scale semantic representations ranging from fine-grained high-frequency details to coarse-grained low-frequency patterns. 
This hierarchical multi-scale semantic understanding is progressively enhanced as we transition from low-resolution to high-resolution components with increasing levels of decomposition, as demonstrated in the reconstruction quality analysis presented in Figure \ref{image:level_reconstruction_quality}.

The experimental results reveal a substantial performance boost when employing 3-4 levels of decomposition compared to 1-2 levels as shown in Table \ref{table:decomposition_level}, clearly demonstrating the critical importance of hierarchical decomposition in learning effective representations for remote sensing applications. 
This improvement can be attributed to the model's enhanced capacity to capture multi-scale spatial patterns and frequency-domain relationships that are characteristic of Earth observation imagery, where meaningful semantic information spans multiple resolution scales from local texture details to regional landscape structures.
We did not extend the decomposition to a fifth level, as the increased memory requirements would exceed the available hardware capacity in the configuration considered. 
Moreover, preliminary tests indicated diminishing performance improvements beyond the fourth level, making four levels a practical and computationally efficient choice.

\begin{table*}[t]
  \centering
  \caption{Ablation study on WaveMAE method applying an increasing level of decomposition of the DWT. Each model has been tested across five datasets. Results reported as mean $\pm$ standard deviation over 5 runs with different seed. Best result per column is \textbf{bold}, second best is \underline{underlined}.}
  \label{table:decomposition_level}
  \resizebox{\textwidth}{!}{%
  \begin{threeparttable}
        \begin{tabular}{@{} l c c c c c c @{}}
            \toprule
            \makecell{Decomposition\\level} & \makecell{\#Params \\(Encoder only)} &
            \makecell{Mados \\mIoU $\uparrow$ $\pm$ std} &
            \makecell{HLSBurnScars \\mIoU $\uparrow$ $\pm$ std} &
            \makecell{BioMassters \\mRMSE $\downarrow$ $\pm$ std} &
            \makecell{Spacenet7 \\mIoU $\uparrow$ $\pm$ std} &
            \makecell{BigEarthNet \\mAP $\uparrow$ $\pm$ std} \\
            \midrule
            %v6_2
            \makecell{1 Level} & 85.21M & 43.421 $\pm$ 1.543 & 83.123 $\pm$ 0.726 & 125.107 $\pm$ 0.077 & 42.633 $\pm$ 3.787 & 42.710 $\pm$ 0.423 \\
            %v6_9
            \makecell{2 Levels} & 85.32M & 43.323 $\pm$ 0.849 & 83.272 $\pm$ 0.664 & 124.167 $\pm$ 0.334 & 42.222 $\pm$ 1.662 & 43.206 $\pm$ 0.211 \\
            %v6_10
            \makecell{3 Levels} & 85.43M & \underline{46.840} $\pm$ \underline{1.980} & \textbf{84.703} $\pm$ \textbf{0.291} & \underline{123.741} $\pm$ \underline{0.151} & \underline{44.582} $\pm$ \underline{1.981} & \textbf{44.224 $\pm$ 0.468} \\
            %v6
            \makecell{4 Levels} & 85.55M & \textbf{50.401 $\pm$ 2.963} & \underline{83.294} $\pm$ \underline{0.918} & \textbf{123.244 $\pm$ 0.122} & \textbf{48.280 $\pm$ 2.292} & \underline{43.760} $\pm$ \underline{0.657} \\
            \midrule
        \end{tabular}
  \end{threeparttable}
  }%
\end{table*}

\subsubsection{Geo-conditioned Positional Encoding}
\label{subsec:gpe_ablation}

We assess the impact of Geo-conditioned Positional Encoding (GPE) both in terms of intrinsic alignment between representations and geographical distance, as well as on downstream transfer performance.
GPE injects latitude and longitude priors, encoded via Spherical Harmonics, into the encoder, thereby encouraging embeddings of geographically close samples to reside nearer in the feature space, while maintaining semantic consistency.
As a result, we expect a stronger monotonic correlation between GPE-based encodings of coordinate pairs and their geographical distance, as well as improved separation between geographically near and far samples in the embedding space.
The geographical distance is expressed with the Haversine formula \cite{haversine_paper} defined as:
\begin{equation}
    \begin{aligned}
        \Delta_{hav}(p_1, p_2) = 2R \cdot \arccos ( &\sin \phi_1 \sin \phi_2 +\\
        & \cos \phi_1 \cos \phi_2 \cos(\theta_2 - \theta_1))\\
    \end{aligned}
\end{equation}
where $R$ is the Earth’s radius ($R=6371$km) and $p_1, p_2$ are two points expressed in radians where $\theta$ is the latitude and $\phi$ is the longitude. 
The Haversine formula determines the great-circle distance between two points on a sphere given their longitudes and latitudes.
Such implicit separation, when combined with the component loss $\mathcal{L}_{cmp}$, is expected to yield consistent gains in downstream performance.

\textbf{GPE-pairs.}
First, we want to prove that GPE carries geolocation information, thus we show the correlation between distance between two pair of coordinates, computed by the Haversine formula $\Delta_{hav}$, and the similarity of learned GPE encodings using the same pair of coordinates.
We sampled $\sim5$k pairs of coordinates $(p_i, p_j)$, with $p = (\phi,\theta) = (\text{latitude}, \text{longitude})$, from images of fMoW-S2 equally distributed between close pairs ($<200\,$ km distance between points with a mean distance of $75$km) or far pairs ($>200\,$ km distance between points with a mean distance of $6965$km).

For each point $p$ we compute also the associated GPE.
For each pair $(p_i, p_j)$ we compute the cosine similarity between their associated GPEs and calculate the Spearman correlation \cite{spearman_paper} between the latter and their great-circle distance $\Delta_{hav}(p_i, p_j)$.
The Spearman correlation coefficient $r_s$ measures the strength and direction of a monotonic relationship and is calculated by the following formula:
\begin{equation}
r_s = 
\frac{
    \displaystyle \sum_{i=1}^{n} (R_i - \bar{R})(S_i - \bar{S})
}{
    \sqrt{\displaystyle \sum_{i=1}^{n} (R_i - \bar{R})^2}
    \sqrt{\displaystyle \sum_{i=1}^{n} (S_i - \bar{S})^2}
}
\end{equation}
where $R_i$ and $S_i$ denote the ranks of the $i$-th observations in the two variables (i.e., Haversine distance and cosine similarity), 
$\bar{R}$ and $\bar{S}$ are their mean ranks, and $n$ is the number of paired samples. 
The coefficient $r_s$ varies between -1 and +1 with 0 implying no correlation, while correlations of -1 or +1 imply a strong monotonic relationship.
We evaluate the correlation between GPE-pairs in all configurations using GPE: GPE only, and GPE+$\mathcal{L}_{cmp}$.

Table \ref{table:gpe_pairs} shows that in both cases the cosine distances between GPEs far and close pairs are pronounced, creating a distinct separation.
Moreover, in both cases, the negative Spearman coefficient $r_s$ close to -1 indicates an inverse strong monotonic relationship between the GPE pairs and the $\Delta_{hav}$.
This implies that geographically distant points on the globe produce GPEs with lower similarity, whereas geographically close points yield encodings with higher cosine similarity, thus confirming our initial hypothesis.

\begin{table}[htb]
    \centering
    \caption{Intrinsic GPE evaluation on GPE-pairs using a frozen encoder. 
    Reported are mean cosine similarity for close/far pairs ($Sim_{cos}$) and Spearman correlation coefficient ($r_s$) between cosine similarity and $\Delta_{hav}$ (km).}
    \label{table:gpe_pairs}
    \resizebox{\columnwidth}{!}{%
    \begin{threeparttable}
    \begin{tabular}{@{} c c c c c @{}}
        \toprule
        \makecell{GPE} &
        \makecell{$\mathcal{L}_{cmp}$} &
        \makecell{Close pairs\\(mean sim. $\pm$ std) $\uparrow$} &
        \makecell{Far pairs\\(mean sim. $\pm$ std) $\downarrow$} &
        \makecell{Spearman corr.\\($r_s$)} \\
        \midrule
        \checkmark &  & 0.992 $\pm$ 0.011 & 0.401 $\pm$ 0.22 & -0.931 \\
        \checkmark & \checkmark & 0.995 $\pm$ 0.007 & 0.592 $\pm$ 0.202 & -0.928 \\
        \bottomrule
    \end{tabular}
    \end{threeparttable}
    }%
\end{table}

\begin{table}[htb]
    \centering
    \caption{Intrinsic GPE evaluation on Geo-conditioned embeddings using a frozen encoder. 
    Reported are mean cosine distances between anchor (A) and positives/negatives.}
    \label{table:gpe_embeddings}
    \resizebox{\columnwidth}{!}{%
    \begin{threeparttable}
    \begin{tabular}{@{} c c c c c c c @{}}
        \toprule
        \makecell{GPE} &
        \makecell{$\mathcal{L}_{cmp}$} &
        \makecell{(A,P)\\$dist_{cos}$ $\downarrow$} &
        \makecell{(A,N)\\$dist_{cos}$ $\downarrow$} &
        \makecell{(A,EP)\\$dist_{cos}$ $\uparrow$} &
        \makecell{(A,EN)\\$dist_{cos}$ $\uparrow$} &
        \makecell{Margin (N,P)\\$\Delta_{AN\rightarrow AP}$}\\
        \midrule
        &  & 0.119 & 0.138 & 0.084 & 0.169 &  0.019\\
        & \checkmark & 0.101 & 0.106 & 0.072 & 0.123 & 0.005\\
        \checkmark &  & 0.111 & 0.137 & 0.079 & 0.163 & 0.026\\
        \checkmark & \checkmark & 0.115 & 0.128 & 0.082 & 0.148 & 0.013\\
        \bottomrule
    \end{tabular}
    \end{threeparttable}
    }%
\end{table}

\textbf{GPE-embeddings.}
Second, in the Geo-conditioned embedding evaluation, we want to evaluate how GPE influences the representations space compared to configurations that do not incorporate it.
To accomplish this, we construct a balanced set of $\sim3$k tuples across the 62 fMoW-S2 categories. 
For each tuple, we define an Anchor (A), that is an image with associated metadata sampled randomly from one of the categories of the dataset.
For each Anchor we concatenate: 
\begin{itemize}
    \item a Positive (P) image with $\Delta_{hav}(A,P)<200$\,km with a different category $c_A \neq c_P$, 
    \item an Easy Positive (EP) image with $\Delta_{hav}(A,EP)<200$\,km with the same category $c_A \equiv c_P$, 
    \item a Negative (N) image with $\Delta_{hav}(A,N)>200$\,km with the same category $c_A \equiv c_P$, 
     \item and an Easy Negative (EN) image with $\Delta_{hav}(A,EN)>200$\,km with a different category $c_A \neq c_P$.
\end{itemize}
For each element in the tuple (A, P, EP, N, EN) we compute the embeddings using our pretrained encoder in 4 different configurations: no GPE / No $\mathcal{L}_{cmp}$, GPE only, $\mathcal{L}_{cmp}$ only, and GPE+$\mathcal{L}_{cmp}$.
We compute the average cosine distance between the Anchor and all other embeddings in the tuple.
Results are shown in Table \ref{table:gpe_embeddings}. 
All intrinsic tests employ a frozen encoder from each configuration considered, to isolate representational effects.

\textbf{Insights.}
As previously demonstrated, GPE induces a strong negative correlation between cosine similarity and geographical distance, with consistently high similarity for geographically close pairs of locations and substantially reduced similarity for distant ones. 
Comparing the two configurations without GPE in Table \ref{table:gpe}, we observe that the best downstream performance corresponds to smaller margins between Positive and Negative embeddings, as highlighted in Table \ref{table:gpe_embeddings}. 
This suggests that a more homogeneous representation space may be advantageous for transfer learning.
Conversely, the strong bias introduced by GPE induces the largest separation between Positives and Negatives (Table \ref{table:gpe_embeddings}), this separation however does not translate into improved downstream performance, rather the opposite, as shown in Table \ref{table:gpe}. 
We speculate that the unregulated shift of tokens in the embedding space caused by GPE is excessively strong, impairing the model’s reconstruction ability and thus degrading the learned representations.
This hint is the reason behind the introduction of a regularization loss to enforce reconstruction consistency.
By introducing the loss $\mathcal{L}_{cmp}$ we mitigated the overly strong bias introduced by GPE while retaining its semantics, thereby allowing the injection of geolocation priors without excessively disrupting the latent space. 
Although the resulting margin remains slightly larger than that of $\mathcal{L}_{cmp}$ only configuration, the embedded geographic priors yield significant improvements in downstream tasks across segmentation, regression, and change-detection tasks, as summarized in Table~\ref{table:gpe}. 

\begin{table*}[tbh]
    \centering
    \caption{Ablation on the effect of GPE and $\mathcal{L}{cmp}$ on downstream tasks. Results reported as mean $\pm$ standard deviation over 5 runs with different seed. Best result per column is \textbf{bold}, second best is \underline{underlined}.}
    \label{table:gpe}
    \resizebox{\textwidth}{!}{%
    \begin{threeparttable}
    \begin{tabular}{@{} c c c c c c c c @{}}
        \toprule
        \makecell{GPE} & 
        \makecell{$\mathcal{L}_{cmp}$} & 
        \makecell{\#Params \\(Encoder only)} &
        \makecell{Mados \\mIoU $\uparrow$ $\pm$ std} &
        \makecell{HLSBurnScars \\mIoU $\uparrow$ $\pm$ std} &
        \makecell{BioMassters \\mRMSE $\downarrow$ $\pm$ std} &
        \makecell{Spacenet7 \\mIoU $\uparrow$ $\pm$ std} &
        \makecell{BigEarthNet \\mAP $\uparrow$ $\pm$ std} \\
        \midrule

        % v6
        & & 85.55M & \underline{50.401} $\pm$ \underline{2.963} & \underline{83.294} $\pm$ \underline{0.918} & 123.244 $\pm$ 0.122 & 48.280 $\pm$ 2.292 & 43.760 $\pm$ 0.657 \\
        % v6_11
        & \checkmark & 85.55M &  47.694 $\pm$ 1.934 & 83.288 $\pm$ 0.650 & \underline{123.071} $\pm$ \underline{0.168} & \underline{48.574} $\pm$ \underline{2.875} & \textbf{43.838 $\pm$ 0.334} \\
        % v6_1
        \checkmark & & 85.55M & 48.997 $\pm$ 2.272 & 82.705 $\pm$ 1.056 & 123.516 $\pm$ 0.120 & 47.586 $\pm$ 1.779 & 43.364 $\pm$ 0.521 \\
        % v6_13
        \checkmark & \checkmark & 85.55M & \textbf{52.258 $\pm$ 2.549} & \textbf{83.565 $\pm$ 1.353} & \textbf{122.843 $\pm$ 0.173} & \textbf{50.703 $\pm$ 2.216} & \underline{43.830} $\pm$ \underline{0.483} \\
        \midrule
    \end{tabular}
    \end{threeparttable}
    }%
\end{table*}

\subsubsection{Masking ratio}
The masking ratio determines the fraction of tokens removed from the input sequence during pretraining and is thus a critical factor in controlling the difficulty of the reconstruction task. Table~\ref{table:mask_ratio} reports the performance of models trained with different masking ratios (60\%, 75\%, and 90\%) across all downstream datasets.  

We observe that a masking ratio of 75\% achieves the most consistent results, yielding the best or second-best performance across all tasks. 
In particular, it provides strong gains on Mados, BioMassters, and Spacenet7, while maintaining competitive performance on HLSBurnScars and BigEarthNet. 
The 60\% configuration performs well on HLSBurnScars and yields competitive reconstruction on BioMassters, but underperforms on the remaining datasets. Conversely, a very high masking ratio of 90\% shows comparable performance to 75\%, with slightly lower variance but generally weaker results.  

These findings confirm the importance of appropriately tuning the masking ratio: while lower masking ratios make the task too easy, leading to weaker representations, overly aggressive masking (90\%) limits the model’s ability to exploit contextual information. 
In our setting, a 75\% masking ratio emerges as the most balanced choice, encouraging the encoder to learn robust and transferable representations.

\begin{table*}[bth]
    \centering
    \caption{Ablation on masking ratio applied to the sequence in input to the encoder during pretraining. Each model has been tested across five datasets. Results reported as mean $\pm$ standard deviation over 5 independent runs with different seed. Best result per column is \textbf{bold}, second best is \underline{underlined}.}
    \label{table:mask_ratio}
    \resizebox{\textwidth}{!}{%
    \begin{threeparttable}
    \begin{tabular}{@{} l c c c c c c c @{}}
        \toprule
        \makecell{Mask ratio} & 
        \makecell{\#Params \\(Encoder only)} & 
        \makecell{Mados \\mIoU $\uparrow$ $\pm$ std} &
        \makecell{HLSBurnScars \\mIoU $\uparrow$ $\pm$ std} &
        \makecell{BioMassters \\mRMSE $\downarrow$ $\pm$ std} &
        \makecell{Spacenet7 \\mIoU $\uparrow$ $\pm$ std} &
        \makecell{BigEarthNet \\mAP $\uparrow$ $\pm$ std} \\
        \midrule
        % v6_17
        \makecell{60\%} & 85.55M & 44.571 $\pm$ 1.161 & \textbf{83.586 $\pm$ 0.617} & \underline{123.108} $\pm$ \underline{0.100} & 49.061 $\pm$ 2.025 & 43.602 $\pm$ 0.763 \\
        % v6_13
        \makecell{75\%} & 85.55M & \textbf{52.258 $\pm$ 2.549} & \underline{83.565} $\pm$ \underline{1.353} & \textbf{122.843 $\pm$ 0.173} & \textbf{50.703 $\pm$ 2.216} & \textbf{43.830 $\pm$ 0.483} \\
        % v6_18
        \makecell{90\%} & 85.55M & \underline{51.266} $\pm$ \underline{0.660} & 83.540 $\pm$ 0.249 & 123.162 $\pm$ 0.149 & \underline{50.108} $\pm$ \underline{2.168} & \underline{43.818} $\pm$ \underline{0.602} \\
        \midrule
    \end{tabular}
    \end{threeparttable}
  }%
\end{table*}

\subsubsection{Pretrain effectiveness}  
We next assess the contribution of pretraining compared to a randomly initialized encoder. 
Table~\ref{table:pretrain_effectiveness} reports downstream results for the full WaveMAE architecture pretrained on fMoW-S2 versus training from scratch with identical settings. 

Pretraining provides a clear and consistent advantage across almost all benchmarks. In particular, it yields large gains on Mados ($+11.4$ mIoU), Spacenet7 ($+8.5$ mIoU), and BioMassters ($-1.46$ mRMSE), highlighting the effectiveness of the learned representations in segmentation and regression tasks. Improvements on \textit{HLSBurnScars} are more modest but still positive, while \textit{BigEarthNet} shows parity with random initialization, suggesting that large-scale pretraining is less critical for multi-label classification when training data is abundant.

These results confirm that pretraining on large, diverse optical datasets such as \textit{fMoW-S2} substantially boosts transfer performance, especially on tasks with limited supervision. 
Random initialization struggles to match this generalization, underscoring the importance of pretraining for representation quality and downstream robustness.  

\begin{table*}[bth]
    \centering
    \caption{Ablation on pretrain effectiveness against a random initialization considering the full WaveMAE architecture. Each model has been tested across five datasets. Results reported as mean $\pm$ standard deviation over 5 independent runs with different seed. Best result per column is \textbf{bold}.}
    \label{table:pretrain_effectiveness}
    \resizebox{\textwidth}{!}{%
    \begin{threeparttable}
    \begin{tabular}{@{} l c c c c c c @{}}
        \toprule
        \makecell{Pretrain Dataset} & 
        \makecell{\#Params \\(Encoder only)} & 
        \makecell{Mados \\mIoU $\uparrow$ $\pm$ std} &
        \makecell{HLSBurnScars \\mIoU $\uparrow$ $\pm$ std} &
        \makecell{BioMassters \\mRMSE $\downarrow$ $\pm$ std} &
        \makecell{Spacenet7 \\mIoU $\uparrow$ $\pm$ std} &
        \makecell{BigEarthNet \\mAP $\uparrow$ $\pm$ std} \\
        \midrule
        % v6_13_scratch
        \makecell{Random Init.} & 85.55M & 40.816 $\pm$ 1.560 & 81.932 $\pm$ 0.838 & 124.303 $\pm$ 0.097 & 42.215 $\pm$ 1.503 & \textbf{43.842} $\pm$ \textbf{0.429} \\
        % v6_13
        \makecell{fMoW-S2} & 85.55M &\textbf{52.258 $\pm$ 2.549} & \textbf{83.565 $\pm$ 1.353} & \textbf{122.843 $\pm$ 0.173} & \textbf{50.703 $\pm$ 2.216} & 43.830 $\pm$ 0.483 \\
        \midrule
    \end{tabular}
    \end{threeparttable}
    }%
\end{table*}

\subsubsection{Token size}
\label{subsec:token_size}

We finally ablate the spatial size of the tokens used as input to the encoder. Table~\ref{table:mask_ratio} compares models pretrained with $16 \times 16$ versus $8 \times 8$ patching, keeping the overall architecture unchanged.  
The results clearly demonstrate that finer tokenization substantially improves downstream performance across all tasks. 

These findings highlight the importance of preserving fine-grained spatial information in the encoder input. Smaller tokens allow the model to capture higher-frequency details and subtle local variations, which appear crucial for both dense prediction (segmentation, regression) and global classification tasks. 
The additional computational cost from increasing the token count is modest compared to the downstream benefits, suggesting that $8 \times 8$ patching offers a favorable trade-off for remote sensing representation learning.

\begin{table*}[bth]
    \centering
    \caption{Ablation on token spatial size, results shows that increasing the granularity of the input tokens leads to better performance overall. Each model has been tested across five datasets. Results reported as mean $\pm$ standard deviation over 5 runs with different seed. Best result per column is \textbf{bold}.}
    \label{table:token_size}
    \resizebox{\textwidth}{!}{%
    \begin{threeparttable}
    \begin{tabular}{@{} c c c c c c c @{}}
        \toprule
        \makecell{Patch size} & 
        \makecell{\#Params \\(Encoder only)} & 
        \makecell{Mados \\mIoU $\uparrow$ $\pm$ std} &
        \makecell{HLSBurnScars \\mIoU $\uparrow$ $\pm$ std} &
        \makecell{BioMassters \\mRMSE $\downarrow$ $\pm$ std} &
        \makecell{Spacenet7 \\mIoU $\uparrow$ $\pm$ std} &
        \makecell{BigEarthNet \\mAP $\uparrow$ $\pm$ std} \\
        \midrule
        % v6_13
        16 & 85.55M & 52.258 $\pm$ 2.549 & 83.565 $\pm$ 1.353 & 122.843 $\pm$ 0.173 & 50.703 $\pm$ 2.216 & 43.830 $\pm$ 0.483 \\
        % v6_15
        8 & 85.91M & \textbf{64.550} $\pm$ \textbf{3.307} & \textbf{85.595} $\pm$ \textbf{0.198} & \textbf{106.383} $\pm$ \textbf{0.170} & \textbf{52.303} $\pm$ \textbf{1.987} & \textbf{45.278} $\pm$ \textbf{0.376} \\
        \midrule
    \end{tabular}
    \end{threeparttable}
    }%
\end{table*}

\subsection{Comparison with State-of-the-Art}
\label{subsec:sota}

At last, we compare WaveMAE against representative state-of-the-art models for optical remote sensing representation learning, namely MAE~\cite{mae_paper}, SatMAE~\cite{satmae_paper}, SatMAE++~\cite{satmae++_paper}, and SpectralGPT~\cite{spectralgpt_paper}.
All methods are pretrained on the same dataset with comparable configurations, fMoW-S2~\cite{fmow_paper}, ensuring a fair and controlled evaluation protocol. 
Table~\ref{table:final_comparison} reports results across the five downstream benchmarks considered in this study.
As previously discussed, to ensure a fair comparison between WaveMAE and competing methods, we adopt an input resolution of $224 \times 224 \times C$ with a patch size of $16$. 
This configuration, in WaveMAE, corresponds to the same effective ratio as an input component resolution of $112 \times 112 \times C$ with a base patch size of $8$ in the first level of decomposition.

\begin{figure*}[!htp]
\centerline{\includegraphics[width=\linewidth]{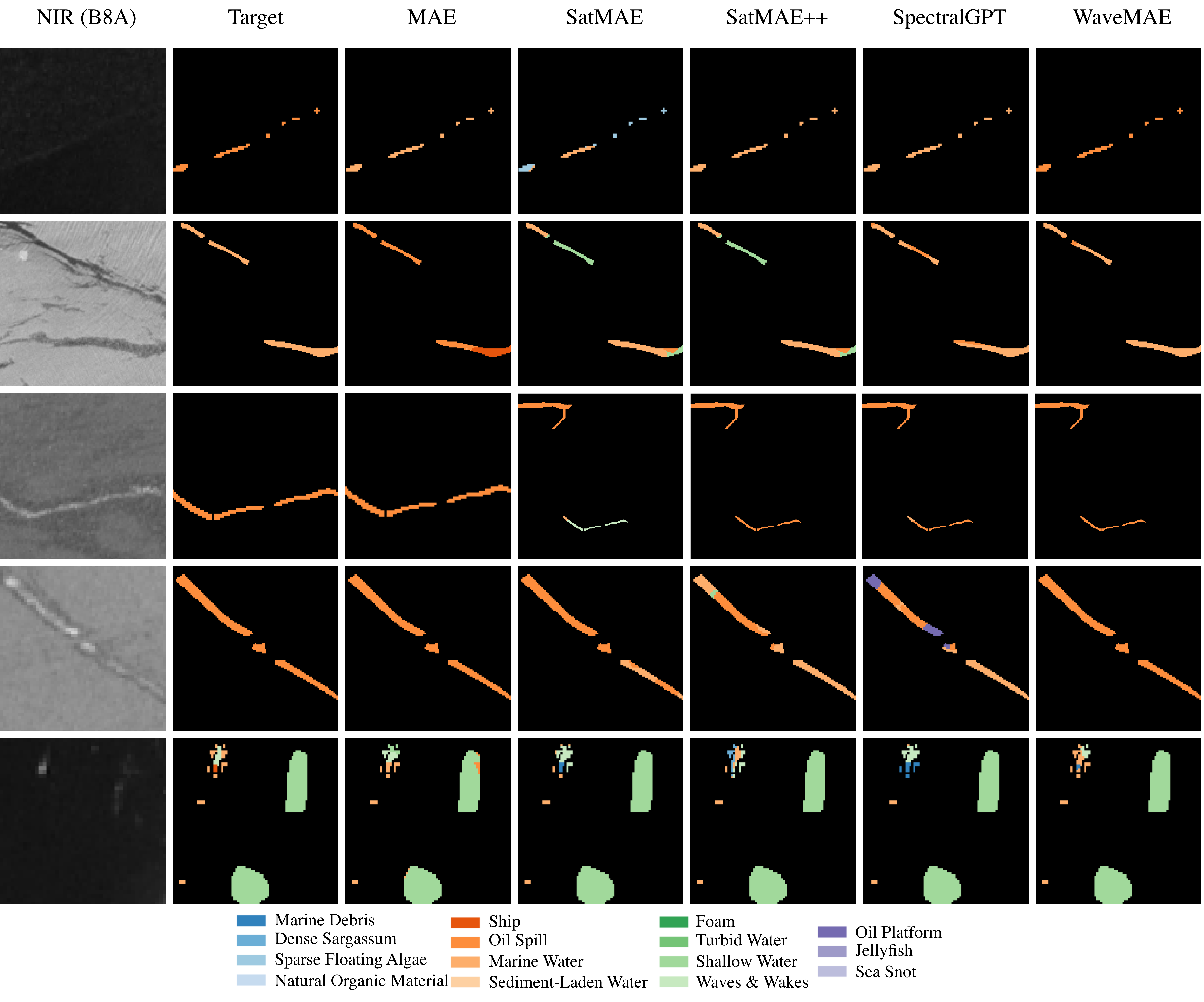}}
\caption{
    Qualitative comparison of semantic segmentation task performance on Mados dataset across different SSL methods. 
    From left to right we show the NIR band (B8A) of the input image, the target segmentation map, and following the predicted segmentation maps respectively produced by MAE \cite{mae_paper}, SatMAE \cite{satmae_paper}, SatMAE++ \cite{satmae++_paper}, SpectralGPT \cite{spectralgpt_paper} and finally our \textbf{WaveMAE}. 
    The labels on the bottom indicate the dataset's classes and the respective color assigned in the segmentation map.
    (Best viewed zoomed in for fine details.)
}
\label{fig:qualitative_mados}
\end{figure*}

\begin{figure*}[!htp]
\centerline{\includegraphics[width=\linewidth]{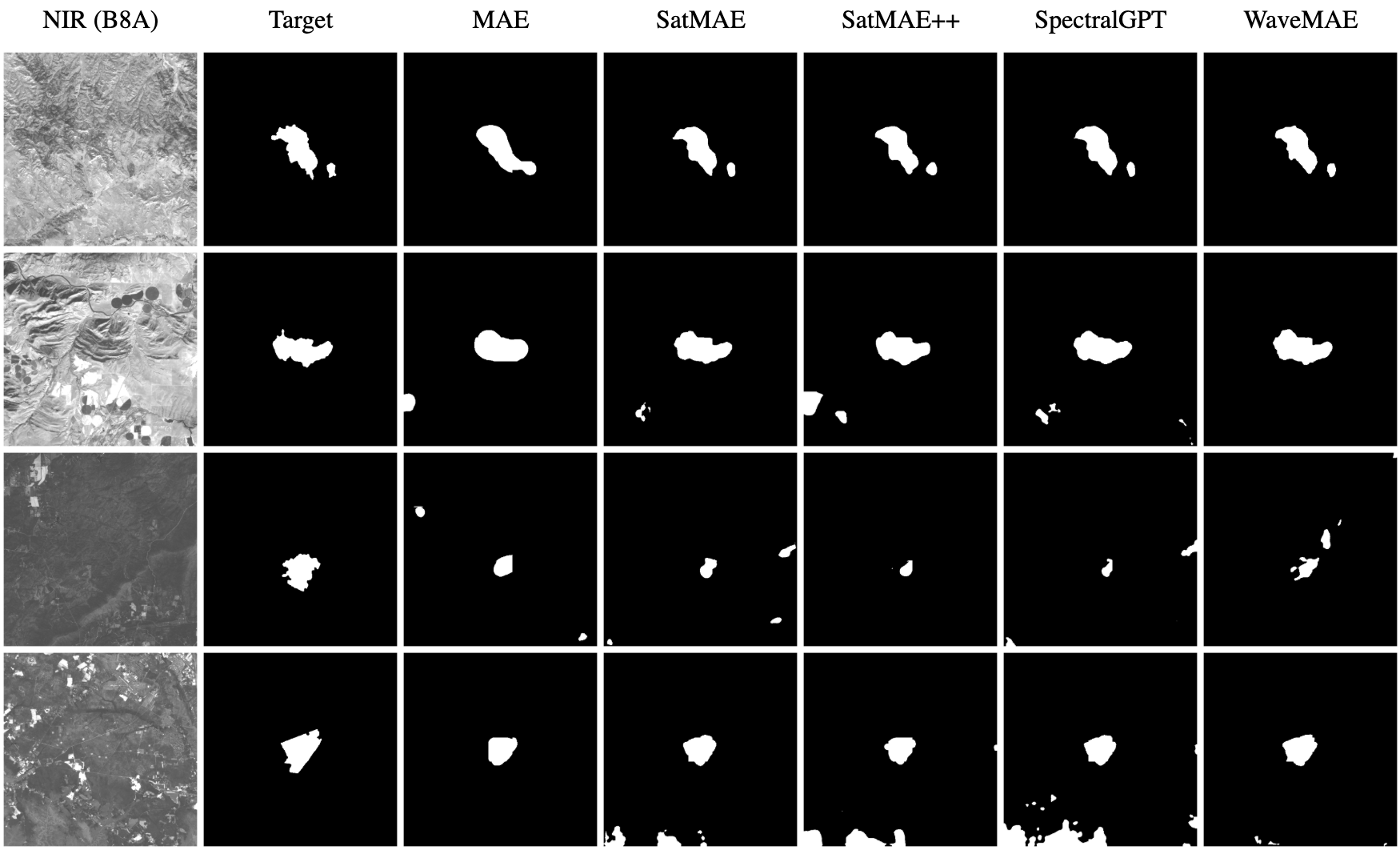}}
\caption{
    Qualitative comparison of semantic segmentation task performance on HLSBurnScars dataset across different SSL methods. 
    From left to right we show the NIR band (B8A) of the input image, the target segmentation map, and following the predicted segmentation maps respectively produced by MAE \cite{mae_paper}, SatMAE \cite{satmae_paper}, SatMAE++ \cite{satmae++_paper}, SpectralGPT \cite{spectralgpt_paper} and finally our \textbf{WaveMAE}. 
    White color is assigned to show the presence of a burn scar in the image.
}
\label{fig:qualitative_hlsburnscars}
\end{figure*}

\begin{figure*}[!htp]
\centerline{\includegraphics[width=\linewidth]{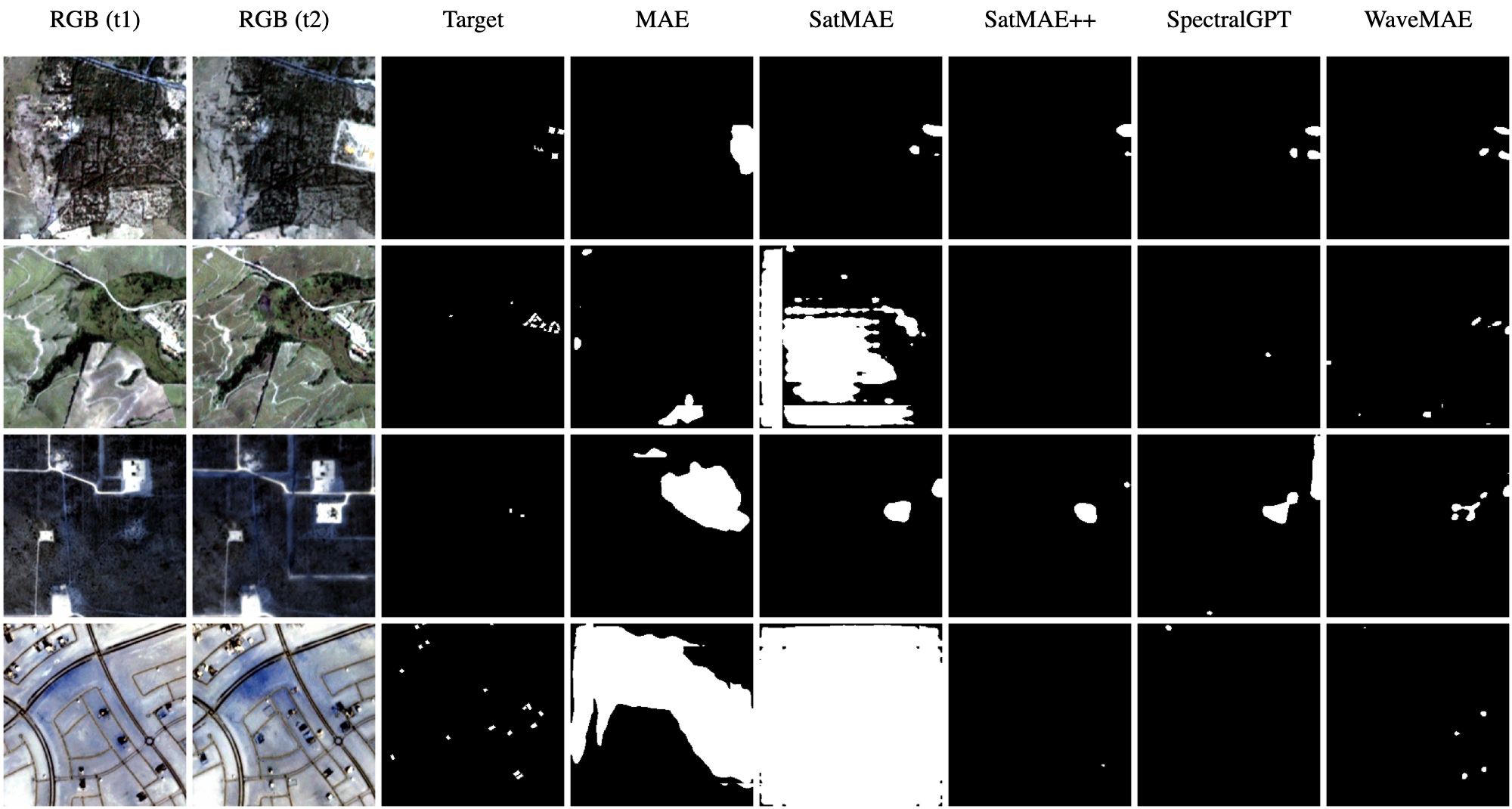}}
\caption{
    Qualitative comparison of change-detection task performance on Spacenet7 (MUDS) dataset across different SSL methods. 
    From left to right we show the RGB image at timestep $t_1$, the RGB image at timestep $t_2$ with $t_1 < t_2$, the target segmentation map, and following the predicted segmentation maps respectively produced by MAE \cite{mae_paper}, SatMAE \cite{satmae_paper}, SatMAE++ \cite{satmae++_paper}, SpectralGPT \cite{spectralgpt_paper} and finally our \textbf{WaveMAE}. 
    The white color in segmentation maps corresponds to areas where urban development occured between the two RGB images.
}
\label{fig:qualitative_spacenet7}
\end{figure*}

WaveMAE, in the base configuration, consistently outperforms prior approaches on four out of five tasks. 
In particular, it achieves substantial improvements on Mados ($+26.2\%$ mIoU) and HLSBurnScars ($+2.35\%$ mIoU), while also setting a new state-of-the-art on BioMassters with a $2.5\%$ reduction in mRMSE. 
For BigEarthNet, WaveMAE surpasses all baselines by a small but consistent margin ($+0.61\%$ mAP).  
The only exception is Spacenet7, where SatMAE++ remains superior, likely due to its enhanced modeling of temporal sequences, which are particularly relevant for change detection tasks.
Moreover, we introduce a WaveMAE-Small variant, which retains performance over previous state-of-the-art while reducing the parameter count to only $26.4$\% of the WaveMAE-Base model.
In this configuration, compared to the Base version, we reduced the encoder depth to 6 (was 12), the embedding size is reduced to 384 (was 768), the number of attention head is 6 (was 12) and the MLP dimension is 1536 (was 3072).
Overall, these results confirm the effectiveness of our wavelet-enhanced masked autoencoding strategy. 
By explicitly encoding scale-aware high-frequency content, WaveMAE produces richer and more transferable representations than existing MAE-based approaches, narrowing the gap between general-purpose pretraining and task-specific architectures in remote sensing.
Finally, a qualitative comparison for all methods considered can be seen in Fig. \ref{fig:qualitative_mados} for semantic segmentation task on Mados dataset, in Fig. \ref{fig:qualitative_hlsburnscars} for semantic segmentation task on HLSBurnScars dataset and in Fig. \ref{fig:qualitative_spacenet7} for change detection task on Spacenet7 dataset.

\begin{table*}[htp!]
    \centering
    \caption{Comparison of WaveMAE against previous state-of-the-art models using fMoW-S2 \cite{fmow_paper} as pretraining dataset for all methods. Each model has been tested across five datasets. Results reported as mean $\pm$ standard deviation over 5 independent runs with different seed. Best result per column is \textbf{bold}, second best is \underline{underlined}.}
    \label{table:final_comparison}
    \resizebox{\textwidth}{!}{%
    \begin{threeparttable}
    \begin{tabular}{@{} l c c c c c c @{}}
        \toprule
        \makecell{Encoder type} & 
        \makecell{\#Params \\(Encoder only)} & 
        \makecell{Mados \\mIoU $\uparrow$ $\pm$ std} &
        \makecell{HLSBurnScars \\mIoU $\uparrow$ $\pm$ std} &
        \makecell{BioMassters \\mRMSE $\downarrow$ $\pm$ std} &
        \makecell{Spacenet7 \\mIoU $\uparrow$ $\pm$ std} &
        \makecell{BigEarthNet \\mAP $\uparrow$ $\pm$ std} \\
        \midrule
        %MAE
        MAE \cite{mae_paper} & 87.76M & 25.414 $\pm$ 2.083 & 83.163 $\pm$ 0.329 & 123.870 $\pm$ 3.887 & 41.416 $\pm$ 1.515 & 45.000 $\pm$ 0.361 \\
        %SATMAE
        SatMAE \cite{satmae_paper} & 87.03M & 47.091 $\pm$ 3.094 & 83.626 $\pm$ 0.274 & 109.040 $\pm$ 0.212 & 45.115 $\pm$ 1.231 & 44.874 $\pm$ 0.401 \\
        %SATMAE++
        SatMAE++ \cite{satmae++_paper} & 87.9M & 39.842 $\pm$ 2.839 & 81.744 $\pm$ 0.472 & 110.358 $\pm$ 0.273 & \textbf{55.228 $\pm$ 1.126} & 43.096 $\pm$ 0.290 \\
        %SPECTRALGPT
        SpectralGPT \cite{spectralgpt_paper} & 85.8M & 51.143 $\pm$ 0.926 & 81.734 $\pm$ 0.822 & 109.201 $\pm$ 0.295 & 49.100 $\pm$ 2.893 & 44.256 $\pm$ 0.145 \\
        \midrule
        %v6_20 - TODO: accennare nel paragrafo di comparison parametri di configurazione aggiuntivi
        WaveMAE-Small & 22.7M & \underline{61.665} $\pm$ \underline{3.422} & \underline{84.063} $\pm$ \underline{1.009} & \underline{108.008} $\pm$ \underline{0.146} & 51.227 $\pm$ 3.770 & \underline{45.126} $\pm$ \underline{0.444} \\
        % v6_15
        WaveMAE-Base & 85.91M & \textbf{64.550 $\pm$ 3.307}  & \textbf{85.595 $\pm$ 0.198} & \textbf{106.383 $\pm$ 0.170} & \underline{52.303} $\pm$ \underline{1.987} & \textbf{45.278 $\pm$ 0.376} \\
        & & (+26.21\%) & (+2.35\%) & (+2.5\%) & (-5.29\%) & (+0.61\%)\\
        \bottomrule
    \end{tabular}
    \end{threeparttable}
    }%
\end{table*}

\section{Conclusions}
\label{sec:conclusion}

In this work, we introduced WaveMAE, a novel self-supervised framework for multispectral remote sensing imagery that extends masked autoencoding with a Discrete Wavelet Transform decomposition and a Geo-conditioned Positional Encoding (GPE). 
Our approach explicitly disentangles spectral-frequency components at multiple resolutions, enabling more effective reconstruction and richer feature representations. 
Through extensive experiments on the PANGAEA-bench, we demonstrated that WaveMAE consistently outperforms prior state-of-the-art foundation models across several downstream tasks.
The effectiveness of WaveMAE pretraining is further demonstrated by showing that even a lightweight variant, containing $26.4$\% of the parameters, achieves state-of-the-art performance.
Ablation studies confirmed the importance of each design choice. 
Wavelet decomposition proved essential for capturing high-frequency details and improving reconstruction fidelity, while GPE-aligned representations with geographical priors enhance both intrinsic embedding structure and downstream transfer performance. 
Furthermore, our analysis of masking ratio, token size, and pretraining effectiveness highlighted the robustness of WaveMAE to hyperparameter variations and its superiority over random initialization. 
Taken together, these findings suggest that frequency-aware decomposition and geographically informed embeddings are key improvements for optically encode remote sensing imagery in self-supervised learning approaches. 
We envision that WaveMAE can serve as a foundation for future large-scale RS pretraining efforts, with potential extensions toward multimodal integration (e.g., SAR-optical fusion) and temporal modeling. 
By providing a fair and systematic comparison across existing methods, this work also lays the groundwork for more standardized evaluation practices in the field.

\section*{Acknowledgments}
Project ECS\_00000033\_ECOSISTER funded under the National Recovery and Resilience Plan (NRRP), Mission 4 Component 2 Investment 1.5 - funded by the European Union – NextGenerationEU.\\
This research benefits from the High Performance Computing facility of the University of Parma, Italy (HPC.unipr.it).


\bibliographystyle{IEEEtran}
\bibliography{egbib}

\vfill

\end{document}